\documentclass[sigconf]{acmart}

\usepackage{graphics}
\usepackage{multirow}
\usepackage{subcaption}
\usepackage{multirow}
\usepackage{xcolor}
\usepackage{graphicx}
\usepackage{threeparttable}
\usepackage[ruled,linesnumbered]{algorithm2e}
\usepackage{pdfpages}
\usepackage[export]{adjustbox}

\usepackage{amsmath}
\usepackage{ifthen}
\usepackage{xargs}

\DeclareMathOperator*{\argmax}{arg\,max}

\newcommandx{\yaHelper}[2][1=\empty]{%
\ifthenelse{\equal{#1}{\empty}}%
  { \ensuremath{ \scriptstyle{ #2 } } } 
  { \raisebox{ #1 }[0pt][0pt]{ \ensuremath{ \scriptstyle{ #2 } } } }  
}   

\newcommandx{\yrightarrow}[4][1=\empty, 2=\empty, 4=\empty, usedefault=@]{%
  \ifthenelse{\equal{#2}{\empty}}
  { \xrightarrow{ \protect{ \yaHelper[ #4 ]{ #3 } } } } 
  { \xrightarrow[ \protect{ \yaHelper[ #2 ]{ #1 } } ]{ \protect{ \yaHelper[ #4 ]{ #3 } } } } 
}

\AtBeginDocument{%
  \providecommand\BibTeX{{%
    \normalfont B\kern-0.5em{\scshape i\kern-0.25em b}\kern-0.8em\TeX}}}

\copyrightyear{2020}
\acmYear{2020}
\setcopyright{acmcopyright}\acmConference[CIKM '20]{Proceedings of the 29th ACM International Conference on Information and Knowledge Management}{October 19--23, 2020}{Virtual Event, Ireland}
\acmBooktitle{Proceedings of the 29th ACM International Conference on Information and Knowledge Management (CIKM '20), October 19--23, 2020, Virtual Event, Ireland}
\acmPrice{15.00}
\acmDOI{10.1145/3340531.3412721}
\acmISBN{978-1-4503-6859-9/20/10}

\begin{document}
\fancyhead{}

\title{Learning to Infer User Hidden States for Online Sequential Advertising}


\author{Zhaoqing Peng, Junqi Jin}
\affiliation{%
  \institution{Alibaba Group}
}


\author{Lan Luo}
\affiliation{%
  \institution{University of Southern California}
}

\author{Yaodong Yang, Rui Luo}
\affiliation{%
 \institution{University College London}
 }

\author{Jun Wang}
\affiliation{%
  \institution{University College London}
}

\author{Weinan Zhang}
\affiliation{%
  \institution{Shanghai Jiao Tong University}
}

\author{Haiyang Xu}
\affiliation{\institution{Alibaba Group}}

\author{Miao Xu, Chuan Yu}
\affiliation{\institution{Alibaba Group}}

\author{Tiejian Luo}
\affiliation{\institution{Univ. of Chinese Academy of Sciences}}

\author{Han Li, Jian Xu, Kun Gai}
\affiliation{%
 \institution{Alibaba Group}
}


\begin{abstract}
To drive purchase in online advertising, 
it is of the advertiser's great interest to optimize the sequential advertising strategy whose performance and interpretability are both important. 
The lack of interpretability in existing deep reinforcement learning methods makes it not easy to understand, diagnose and further optimize the strategy.
In this paper, we propose our Deep Intents Sequential Advertising (DISA) method to address these issues. 
The key part of interpretability is to understand a consumer's purchase intent which is, however, unobservable (called hidden states). 
In this paper, we model this intention as a latent variable and formulate the problem as a Partially Observable Markov Decision Process (POMDP) where the underlying intents are inferred based on the observable behaviors. 
Large-scale industrial offline and online experiments demonstrate our method's superior performance over several baselines. 
The inferred hidden states are analyzed, and the results prove the rationality of our inference.
\end{abstract}



\maketitle

\section{Introduction}

\begin{figure}
\centering
\includegraphics[page=2,height=4.5cm]{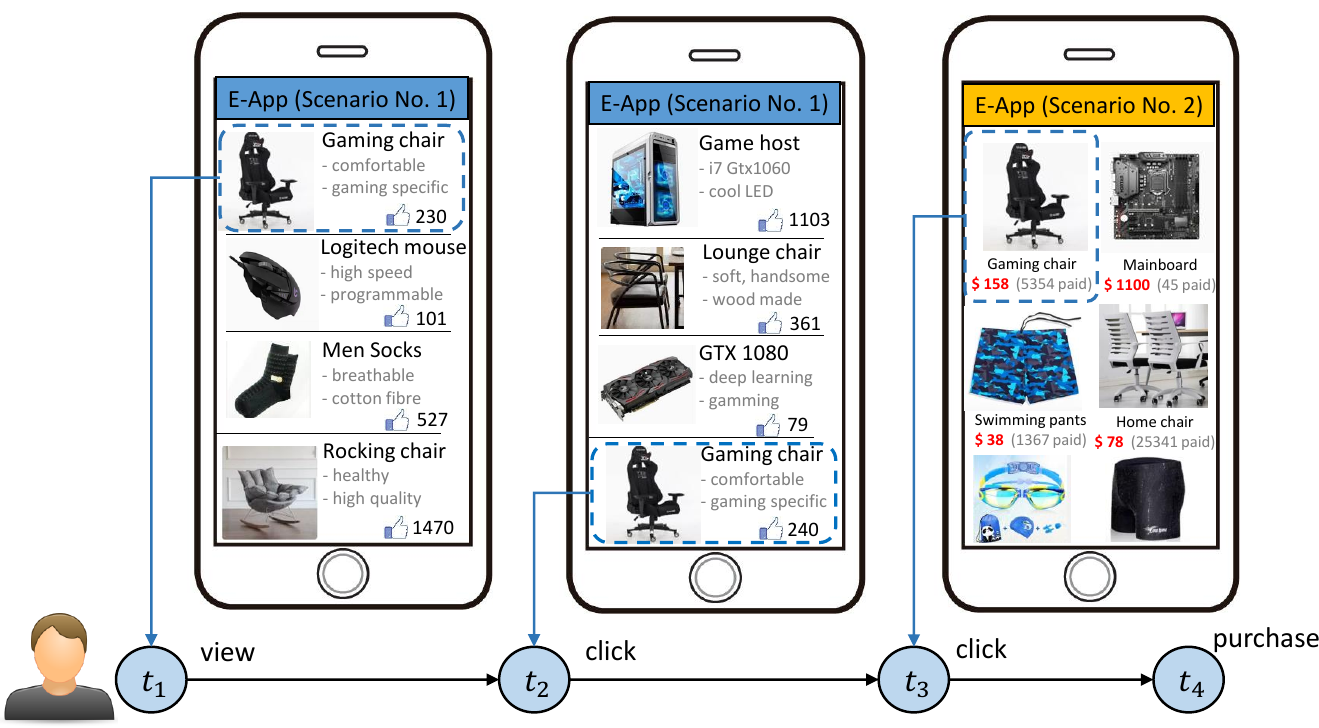}
\hfill
\caption{An advertised item on consumer trajectories across multiple scenarios.}
\label{fig:example}
\end{figure}

Online advertising is an effective way for advertisers to reach their targeted audiences and drive conversions. 
Compared to a single ad exposure, sequential advertising \cite{shao2011data} has a higher chance of cultivating consumers' awareness, interest and driving purchases in several steps through multiple scenarios. 
Fig. \ref{fig:example} shows an example of sequential advertising on a Gaming chair in two scenarios.
At time $t_1$, the consumer browses and becomes aware of the chair in scenario No. 1.
At time $t_2$, he sees it again and shows interest by clicking it. 
After a while, the consumer visits scenario No. 2 and finally clicks and makes a purchase at time $t_3$ and $t_4$. 
To maximize the return on investment (ROI), advertisers have a great desire to optimize sequential advertising strategies.

Advertising strategies' optimization and interpretability are both very crucial. 
The significance of optimization comes from its direct results of the ROI.
Interpretability helps advertisers understand the strategy, provides ways to diagnose, conduct conversion attribution, and finally supports further optimization.

The advertising algorithm design for combining performance and interpretability is very challenging. 
The key to interpretability is modeling the consumer's mental states under a sequence of interactions with ads. 
However, these mental states/intents are difficult to define, and they are even unobservable. 
The only information related is the observed consumer's behaviors, e.g., click and purchase actions. 
Most interpretable algorithms tend to use shallow models such as logistic regression (non-neural network) for more convenient analysis; however, they cannot benefit from current advances of deep learning techniques \cite{Rodriguez, mahmud2010constructing, McCallum}.

To overcome these difficulties, there are several related works. 
Interpretable methods like multi-touch attribution (MTA) \cite{ji2017additional, shao2011data} focus on assigning credits to the previously displayed ads before the conversion, but they usually do not provide future strategy optimization. 
Performance-oriented methods such as deep reinforcement learning (DRL) usually aggregate the consumer's historical behaviors as an input of a black-box neural network and obtain the advertising action directly from the output of the network \cite{cai2017real, jin2018real, feng2018learning, hu2018reinforcement, chen2018stabilizing}. 
This kind of straightforward aggregation of behaviors cannot represent and interpret the consumer's mental states well, which makes understanding, diagnosing, and optimizing the strategy difficult. 
Some algorithms give considerations to both interpretability and strategy optimization \cite{mahmud2010constructing, Murphy, McCallum, mccallum1993overcoming}. 
Nonetheless, the majority of these methods are limited to theoretical analysis, and the experiments are conducted mostly in toy simulated environments, which are impractical for realistic industrial applications. 

Considering the above challenges and shortcomings, we propose our Deep Intents Sequential Advertising (DISA) algorithm to address these issues in advertising applications. 
We formulate the multi-step advertising problem as a Markov Decision Process (MDP). 
In this MDP, the consumer intents (state) are not directly observable, so we use POMDP to model the state as a hidden variable inferred by observed behaviors.
However, as a probabilistic framework, POMDP's parameters are not off-the-shelf. 
To tackle this issue, we derive an expectation–maximization (EM) algorithm to estimate the parameters by learning from large-scale real-world data. 
The learned POMDP model can infer the probability distribution of user hidden states, defined as beliefs. 
Unlike noisy behavior data, beliefs are more abstract, and we can interpret how probable a user visits each hidden state and to which state it may transit. 
Finally, we optimize the sequential advertising strategy depending on the beliefs. 
Since the learning of the exact POMDP's optimum policy is intractable \cite{Murphy}, we approximate the belief value function using a variant of Smooth Partially Observable Value Approximation (SPOVA) method \cite{parr1995approximating}.
It is a more suitable deep architecture for POMDP than pure black-box Deep Q-Network (DQN). 

Offline experiments show that our method is better than several baselines.
Online results demonstrate our sequential advertising's superior performance over the existing system. 
In terms of interpretability, we analyze the inferred hidden states and provide examples of state transitions under different advertising strategies. 

Our main contributions include: 
1) To our best knowledge, our DISA is the first attempt focusing on the interpretability of realistic advertising strategies with POMDP. 
2) To optimize the strategy performance, we propose a variant of SPOVA method, a more suitable deep neural network solution for POMDP than pure black-box deep networks commonly used in DRL. 
3) We develop POMDP's application in large-scale industrial settings. The inferred hidden states are analyzed to show the efficacy of our method.

The rest of this paper is organized as follows. 
Section 2 introduces the recent work related to POMDPs, followed by an analysis of the sequential advertising problem in section 3.
Section 4 formulates the problem.
Section 5 presents our approach to the problem and gives detailed implementations. 
In Section 6, we discuss the experimental results and interpret the hidden states as well as the learned advertising strategies.
Section 7 concludes the paper.

\section{Related Work}

Generally, a POMDP model can be considered as a belief-state MDP \cite{Murphy}. 
The Hidden Markov Model (HMM) is usually used to represent the hidden states of POMDP \cite{McCallum, mccallum1993overcoming, mahmud2010constructing}. 
The Baum-Welch algorithm \cite{Koenig} is extended to adjust the probabilities of the Markov model, while Bayesian-based methods \cite{Ross, Rodriguez} can improve the model through interaction with the environment. 
For policy learning, structured representations are usually used to solve the value approximation \cite{roy2005finding, boutilier1996computing}.
The neural network is first introduced to yield good value approximations in SPOVA \cite{parr1995approximating}, and the recurrent neural network (RNN) is adopted in QMDP-net \cite{Karkus} for the planning of POMDP. 
However, these methods are usually evaluated with simple tasks and impractical for realistic applications.
Although MTA methods \cite{ji2017additional, shao2011data} know how each exposure contributes to the conversion, they do not model user latent states and cannot support online inference; thus, they cannot directly solve our problem.

In applications of MDP and POMDP, bandit-based models with Thompson sampling are widely used in simple recommendation problems \cite{meshram2016optimal}.  
\citet{yuan2012sequential} propose to utilize the correlation of ads to improve the efficiency of exploration.
These applications haven't taken advantage of current deep learning merits for better performance. 
There are some DRL-based solutions \cite{hu2018reinforcement, feng2018learning, chen2018stabilizing, ie2019reinforcement, zhai2016deepintent} to ranking problems. 
\citet{hu2018reinforcement} propose a policy gradient algorithm to learn an optimal ranking policy by modeling the reward function.
\citet{ie2019reinforcement} optimize the slate-based recommendations based on estimated long-term value. 
These works mainly use end-to-end deep learning methods and are weak in terms of interpretability.
DeepIntent \cite{zhai2016deepintent} models the intents using the attention weights on top of RNN, its black-box learning cannot explicitly model intents' transitions; thus, the sample complexity could be higher without the prior knowledge that user behaviors are generated based on hidden state transitions.

\section{Multi-Scenario Sequential Advertising}

In large mobile E-commerce platforms, e.g., Amazon, eBay, Taobao, there are millions of users visiting different scenarios every day.
The repeated visits of these users allow the platform to help advertisers earn more revenue with appropriate multi-step advertising strategies.
In this paper, we follow the framework of MDP, and we care about the interpretations of the displaying effect on a user purchase intention. 
This interpretability benefit us in 1) attribution: easily interpret the insights of user conversions, 2) optimization: guide the future advertising policy in other similar applications.
However, the user intention is not directly observable, so we model it as a hidden state.
To do this, we formalize this problem as a POMDP where the agent (the advertising engine) learns to maximize advertisers' revenue by inferring the consumers' hidden state.

\section{Problem Definition}

Generally, at each time-step $t$, an advertising campaign starts with a user request $U_t$, which contains the user name, age, and historical behaviors. The handling of the request can be formalized as:
(1) \textbf{Matching stage}, by comparing the relevance of different items w.r.t the user, a candidate ad set $\mathcal{D}_t = \{ I_1,I_2, \dots, I_{|\mathcal{D}_t|} \}$ is recalled using some matching methods like TDM \cite{zhu2018learning}s, and $I_i$ is the $i$-th campaign launched by an advertiser $X_i$. 
(2) \textbf{Sorting stage}, the advertising engine performs a ranking function $f_t$ on the set $\mathcal{D}_t$, and top $K$ items $ \mathcal{L}_K^t(\mathcal{D}_t, f_t)= (I_{(1)},I_{(2)}, \dots,I_{(K)})$ are selected and delivered back to the consumer. Here, $K$ is determined by the type of scenarios.
(3) \textbf{Feedback stage}, for each displayed item $I_{i}$, the advertising engine collects the feedback of the user purchase behavior $y_{t}$ and click behavior $x_{t}$. The advertiser $X_i$ will pay money $bid_{i}$ to the advertising engine if the user clicks ($x_{t} =1$) and will obtain revenue $price_{i}$ when the user purchases ($y_{t} =1$).

Formally, given a sequence of requests $\mathcal{Q} = (U_1 \sim U_T)$ from a consumer, our problem is defined as a sequential decision process to determine the appropriate ad items $(\mathcal{L}_K^1 \sim \mathcal{L}_K^T )$\footnote{Usually, the final items $\mathcal{L}_K^t$ can be affected by recalling different ads $\mathcal{D}_t$ in the matching stage or adjusting the ranking function $f_t$ in the sorting stage. In this paper, we only consider how to use $f_t$ to control the final displayed items.} to maximize the advertisers' profits.
The ranking function here is designed to be a set of score actions $f_t = \{a^t_1 \sim a^t_i\}$ on each candidate item $I_i$ in $\mathcal{D}_t$, which are the output of the agent.
To interpret each advertising action $a^t_i$, we need to know how $a^t_i$ will affect or transit a user latent intent\footnote{A user may have multiple intents on different items, and we can feed the model with different items to get different intents.} on the item, which can be explicitly modeled by a POMDP.

Specifically, a POMDP model is a 7-tuple ($\mathcal{S}$, A, O, $T$, $O$, $r$, $\gamma$) where $\mathcal{S}$ is a set of discrete hidden states describing the intents of a user, $A$ is a set of score actions on an item, and $O$ is a set of the agent's observations on user behavior to the item. The transition function $T(s,s',a)=P(s'|s,a)$ describes the probability of transition from state $s$ to $s'$ after executing action $a$, while observation function $O(s',a,o)=P(o|s',a)$ specifies the probability that a next observation $o$ will be received after the agent performs action $a$ and lands in state $s'$. The reward $r$ captures the expected feedback from the environment, and $\gamma \in [0,1]$ is the discounted factor.

\begin{figure}
 \centering
  \includegraphics[height=2.6cm,keepaspectratio]{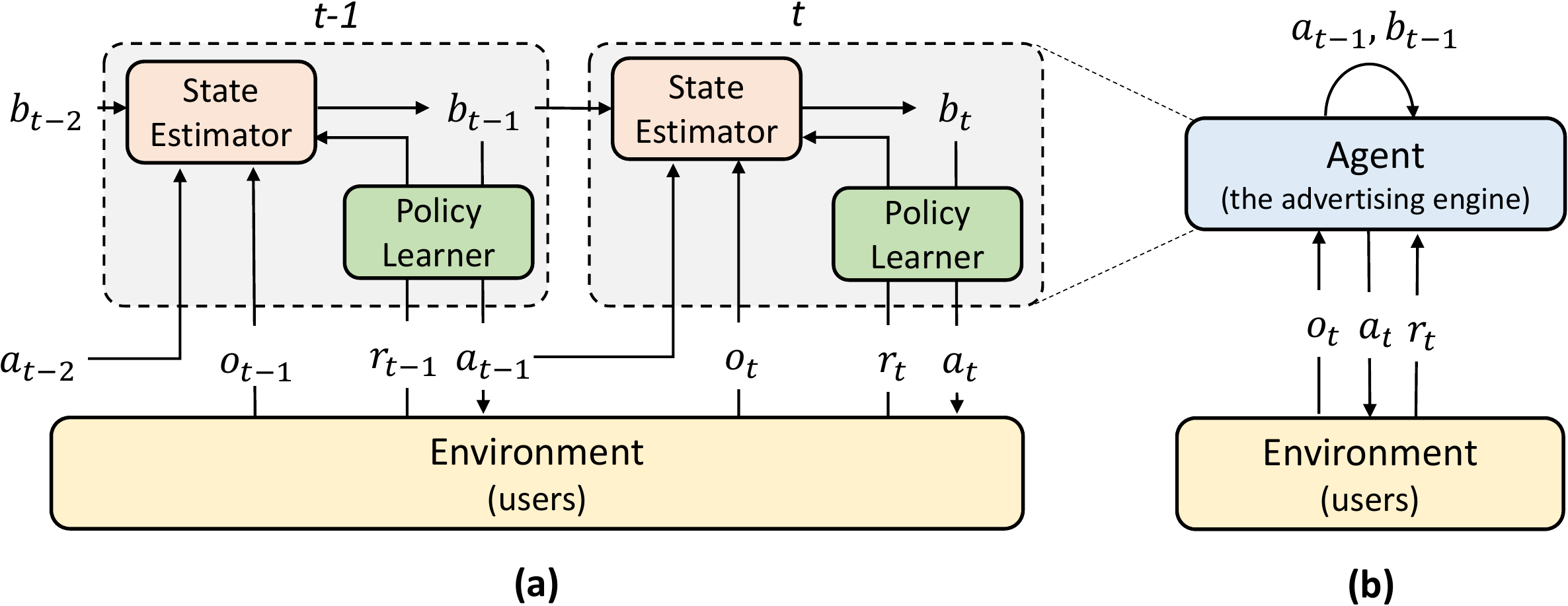}
   \caption{(a) Stream flow unrolled in temporal sequence; (b) the interactions between the agent and the environment.}
   \label{fig:formulation}
\end{figure}

At each time-step $t$, an advertising action $a_t$ is decided given an observation $o_t$, which brings up two steps: 1) the agent infers a belief $b_t$ (defined as a probability distribution over all hidden states) with a state estimator, 2) the action $a_t$ is chosen based on $b_t$ with a policy learner. Fig. \ref{fig:formulation} gives the two steps as following.

\textbf{State Estimator (SE)}. According to the parameters of $T$ and $O$, the state estimator produces the current belief $b_t$ with the observation $o_t$, the previous $b_{t-1}$ and $a_{t-1}$ using the Bayes rule:
\begin{equation} \label{belief updating}
b_{t}(s') = \rho O(s',a_{t-1},o_t)\sum_ {s \in \mathcal{S}} T(s ,s' ,a_{t-1}) b_{t-1}(s),
\end{equation}
where $\rho$ is the normalized factor, and $b(s)$ represents the probability that a consumer hidden state is under state $s$.

\textbf{Policy Learner}. 
After estimating $b_t$, the agent has to learn the mappings from beliefs to actions, denoted by a policy $a_t = \pi(b_{t})$. One could think of a POMDP as an MDP defined over belief states, then the well-known Bellman equation for POMDP still holds \cite{Murphy}. In particular,

\begin{equation} \label{eq1}
\begin{aligned}
V^*(b_t) = \max_{a_t}[ r_t + \gamma \sum_{o \in O}P(o_t|a_t,b_t)V^* (b_{t+1}) ]
\end{aligned} 
\end{equation}
where  $V^*(b_t)$ is the belief value function with an optimal policy $\pi^*$.

Unlike the budget constraint setting in \cite{wu2018budget, jin2018real}, our agent's goal is to maximize the advertiser's profits within a certain time window $T_w$. 
The window is usually set according to how soon most of the conversions are reached after ad exposures. 
The profits are defined as the advertiser's revenue subtracting the budget cost.
The reward is therefore given by $r_{t,i}= price_{i} y_{t}  - bid_{i} x_{t}$. 
The objective of learning is to find an optimal policy to maximize the expected return of each item $I_i$. 
\begin{equation} \label{eq2}
\pi_i^* = \argmax_{\pi_i} \mathbb{E} \left[ \sum_{t=1}^T  \sum_{i \in \mathcal{L}_k(\mathcal{D}_l, f_t)} \gamma^t r_{t,i} | \pi_i  \right]
\end{equation}

\section{Methodology}

In this section, we introduce our proposed DISA with three parts. 
We first present an EM-based method to estimate the parameters of the state estimator. 
We then adopt an approximated method for the policy learner to optimize the value function over beliefs. 
Finally, we give the specific implementation of DISA with the real advertising engine. 

\subsection{EM-based Parameters Estimation}

To perform belief updates with the Eq.~\eqref{belief updating}, we firstly need to know the transition function $T$ and observation function $O$.
However, these two fundamental functions are not available priori in our case, and we have to estimate them in advance.
Essentially, 
a POMDP can be regarded as an extended HMMs conditioned on a sequence of actions.
As such, we can learn the parameters of POMDP by building a conditional HMMs and solving it with EM-based algorithms.

Based on the analysis, we now describe the parameter estimation as a learning problem of a conditional HMM model parameterized by $\theta= (b_0, T, O)$ where $b_0$ is the initial distribution of hidden states.
Given a trajectory $\mathcal{J}_T = (a_0, o_1,a_1, o_2, a_2, \dots, o_T)$ on an ad, we try to find the parameters to best fit the trajectory with user latent variables $\mathcal{S}$. 
Specifically, let $\mathcal{O}=\{o_1 \sim o_{T}\}$ and $\mathcal{A}=\{a_0 \sim a_{T-1}\}$ denote the sequence of observations and corresponding actions in $\mathcal{J}_T $ (each observation is given equal weight), we study the maximization of the log-likelihood of $\mathcal{O}$ conditioned on $\mathcal{A}$:

\begin{equation} \label{eq1}
\begin{aligned}
l(\theta) & = \log P(\mathcal{O} | \mathcal{A};\theta) \\
              & \geq \sum_{s \in \mathcal{S}} q(s) \log P(\mathcal{O},s|\mathcal{A};\theta) - \sum_{s \in \mathcal{S}} q(s) \log q(s) \\
              & = L(q, \theta; \mathcal{O}, \mathcal{A})
\end{aligned}
\end{equation}
where $q(s)$ is a density function that satisfies $\sum_s q(s)=1$. In the lower bound $L(q, \theta; \mathcal{O},\mathcal{A})$, the $\geq$ follows Jensen's inequality, and the equality is only reached at $q(s)=P(s|\mathcal{O},\mathcal{A};\theta)$. 
Following the EM algorithm, at each time-step $t$, our E-step is to estimate:
\begin{equation} \label{eq1}
\begin{aligned}
q^t  = \argmax_q L(q,\theta^{t-1}; \mathcal{O},\mathcal{A}) 
     = P(s|\mathcal{O},\mathcal{A};\theta^{t-1})  
\end{aligned} 
\end{equation}
The M-step is to adjust $\theta$ by maximizing the Q-function with $q^t$:

\begin{equation} \label{eq1}
\begin{aligned}
\theta^t 
             = \argmax_{\theta} E_{q^t(s)} \log P(s,\mathcal{O}|\mathcal{A};\theta^{t-1}) 
\end{aligned} 
\end{equation}

We derive a variant of Baum-Welch algorithm to implement the above iterative procedures, and the details can be found in Supplementary A.1.
When we obtain the estimated $T$ and $O$, the current belief $b_t$ can be updated by Eq.~\eqref{belief updating}.
Our next step is to learn the action policy $a_t=\pi(b_t)$ with the given belief $b_t$.

\subsection{Belief Value Function Approximation}

A critical question for policy learning is how to represent the value function for beliefs. 
Sondik \cite{Rodriguez} showed the value function $V(b)$ can be represented as the max over a finite set of vectors.
However, exact methods for solving this are impractical \cite{Murphy}, and function approximation is a more attractive alternative than exact methods.
In this paper, we prefer to implement the approximation with deep neural networks to improve the learning of our strategies.  
As such, we approximate $V(b)$ using a set of parameterized Q-functions:
\begin{equation} \label{value function}
V(b)=\max_a Q_a(b; \eta_a)  
\end{equation}
where $Q_a(b; \eta_a)$ is the expected return for taking action $a$ in belief $b$, and each Q-function is approximated by a soft max function SPOVA\cite{Rodriguez}: 
\begin{equation} \label{spova}
Q_a(b; \eta_a)= \sqrt[z]{\sum_{i=1}^n (b \cdot \eta_{a_i})^z}  
\end{equation}
here each $\eta_{a_i}$ is the output vector of deep neural networks w.r.t an action $a$, and the value of $n$ determines how many vectors are used to split the belief space into linear representations. 
$z$ is an indicator interpreted as a measure of how "rigid" the approximation is \cite{parr1995approximating}. 
Given the Q-function, our policy $\pi$ is then to select the action with the largest Q-value: $a_t = \argmax_a Q_a(b_t; \eta_a)$.

Assuming $b'$ is the updated belief after performing best action $a$ in $b$, the optimization of the value function is performed by minimizing the square of Bellman residual $E(b)^2$\cite{dqn} where $E(b)=\gamma V(b') +r - Q_a(b; \eta_a)$.
Since Eq.~\eqref{spova} is differentiable, a typical gradient descent method can be used to update each vector $\eta_{a_i}$. 
The updates for the $j$-th component of the $i$-th $\eta$ vector, $\eta_{a_i^j}$ turns out to be:
\begin{equation} \label{updating}
\bigtriangleup \eta_{a_i^j} = \frac {\alpha E(b)b_j(b \cdot \eta_{a_i} )^z} {V(b)^z}
\end{equation}
where $\alpha$ refers to a step size or learning rate. 
Note that we should keep each $\eta_{a_i}$ positive to allow the second derivative of Eq.~\eqref{spova} always positive in each dimension, so the function is always convex. 
This can be done by replacing $(b \cdot \eta_{a})$ with $(b \cdot \eta_{a} + \upsilon)$ where $\upsilon$ is a constant offset \cite{parr1995approximating}.  
However, we found that a large constant $\upsilon$ will bring updating bias when $\gamma > 0$, which leads to an unstable learning process. 
To address this, we compensate the bias in Bellman residual: $E(b)=\gamma V(b') +r - Q_a(b; \eta_a) + (1-\gamma)\upsilon$. 
An alternative is to use reward shaping to keep rewards $r$ always positive, which can prevent the learning direction of $\eta_{a_i}$ from going towards negative values. 

\subsection{Implementations}

Here, we illustrate our detailed solution to the real advertising optimization, including some key concepts of applying DISA, as well as the implementation of the state estimator and policy learner. 

\textbf{Item modeling level}.
From the online data, we found the samples of repeated exposures for a specific item are sparse, which brings difficulties in training.
In this paper, we relax the POMDPs modeling level from items to categories, and different consumers share the parameters of DISA during learning and execution. 
This setting can largely increase the quantity and diversity of learning samples and improve the model's generalization. 
Note that we use the most fine-grained categories maintained by the advertising system. 
According to our data, although the category features will lose some individual information, our categories are detailed enough that the individual differences within a category are small. 
We will study better aggregation methods in the future.

\textbf{Action}. The ranking function $f$ for an advertising platform is usually designed using eCPM sorting mechanism \cite{jin2018real}, which aims to maximize the revenue of the platform, given by $rank\_score = pCTR \times bid$. 
We follow this setting, and we perform actions on the $rank\_ scores$ at the categorical level. 
In particular, we use a ratio $\delta_j$ to adjust the rank score of an item $I_i$ that belongs to the $j$-th category, that is $rank\_score_i' = rank\_score_i \times \delta_j $.
This $\delta$ is the output of the agent's action that could affect the ranking of items so as to decide the final displayed item.
For a discrete action setting, we define three actions: a boosting action with $\delta_j > 1$, a restraining action with $\delta_j < 1$, and a keeping action with $\delta_j = 1$; the value of $\delta$ for each action should be tuned under this setting.

\begin{figure}
\begin{subfigure}{.59\columnwidth}
\centering
\includegraphics[page=5,height=2.5cm]{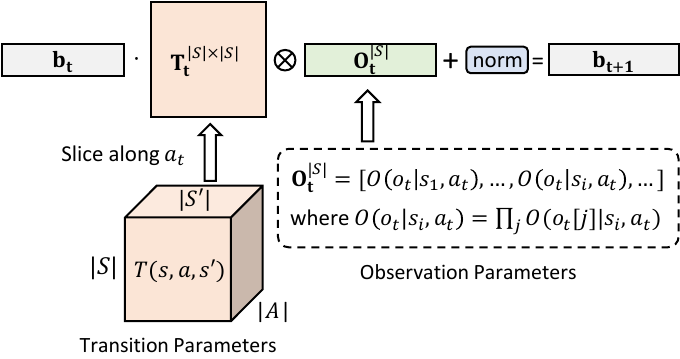}
\caption{Belief updating.}
\end{subfigure}
\begin{subfigure}{.39\columnwidth}
\centering
\includegraphics[page=6,height=2.5cm]{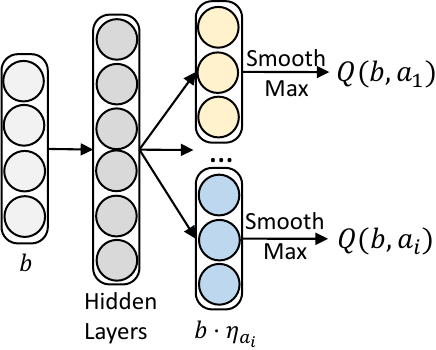}
\caption{Policy network.}
\end{subfigure}
\caption{Implementation of DISA.}
\label{fig:implementation}
\end{figure}

\textbf{Reward}. 
As mentioned in Section 4, our reward is defined as the advertisers' revenue subtracting their budget cost. 
This reward setting may suffer from a local-optimal policy: the agent learns to increase the rewards by reducing advertisers' budget cost. 
To tackle this problem, we propose a bid punishment mechanism to force the agent focusing on improving the revenue rather than reducing budget cost.
We increase the bid price for the boosting action: $bid'= bid \times \beta$ where $\beta \geqslant 1$ is a punishment variable which controls the extra cost of performing the boosting action. Therefore, the reward function is reshaped to be: 
\[
    r_{t,i}= 
\begin{cases}
    \lambda price_{i} y_{t} - \beta_j bid_{i} x_{t},              & \text{if } a_j \text{ is boosting action}\\
    \lambda price_{i} y_{t} - bid_{i} x_{t},              & \text{otherwise}
\end{cases}
\]
where $\lambda$ is set to balance the data magnitude between revenue and cost so that the agent can equally optimize revenue and cost (purchase $y_t$ is more sparse than click $x_t$). 
More details will be discussed in our experiment.

\textbf{State estimator}. 
Following the Eq.~\eqref{belief updating}, we illustrate a belief updating process in Fig. \ref{fig:implementation}(a). 
Considering a case where actions and observations are discrete, the transition function $T: |\mathcal{S}| \times |\mathcal{A}| \times |\mathcal{S}| $ can be parameterized by a 3-dim vector cube. 
Given a $|\mathcal{S}|$-dim belief vector $\mathbf{b_t}$ and a performed action $a_t$, our first step is a dot production: $\mathbf{b_t}=\mathbf{b_t} \cdot \mathbf{T_t}^{|\mathcal{S}| \times |\mathcal{S}|}$ where $\mathbf{T_t}^{|\mathcal{S}| \times |\mathcal{S}|}$ is a transition matrix sliced from $T$ along action $a_t$. 
Suppose there are $G$-dim observations and each dimension is independent with each other, so we have $O(o_t |s_i,a_t) = \prod_{j}^G O(o_t[j] |s_i ,a_t)$, where $O(o_t[j] |s_i ,a_t)$ is the probability of observing $j$-th dimension in $o_t$ given state $s_i$ and action $a_t$. 
Then our second step is an element-wise multiplication of $\mathbf{b_t}$ with a vector $ \mathbf{O_t}^ {|\mathcal{S}|}= [O(o_t|s_1,a_t),O(o_t|s_2,a_t),... ]$, that is $\mathbf{b_t} = \mathbf{b_t} \otimes \mathbf{O_t}^ {|\mathcal{S}|}$. 
Our final step is followed by a normalized operation $\rho$: $ \mathbf{b_t}[i] = \mathbf{b_t}[i]/ \sum_j \mathbf{b_t}[j] $, and it produces the next belief $\mathbf{b_{t+1}}$.

\textbf{Policy learner}. 
The policy learner is implemented with a deep neural network such as multi-layer perception (MLP) as Fig. \ref{fig:implementation}(b) depicts.
The input of this policy network is the belief vector, and the output is split into $|\mathcal{A}|$ groups. 
The output of each group is conducted with the max smooth function of Eq.~ \eqref{spova} to obtain the Q-value function for each action. 
In this case, the $\eta$ vectors for each action are embedded into the parameters of hidden layers, which are trained end-to-end through the whole policy network by a gradient descent method in Eq.~\eqref{updating}.

\textbf{Simulator}.
For offline experiments, we offer a simulator to imitate consumers' feedback by applying supervised learning techniques on real consumer behavior. 
Similar simulator settings can be found in \cite{shi2018virtual, chen2018stabilizing, hu2018reinforcement}. 
Since a user’s preferences can be time-dependent and also depend on the history of past ad impressions, we choose a recurrent model to make multi-task predictions on the real click $x_t$ and purchase $y_t$. 
In particular, at each time-step $t$, we adopt an RNN model to output a vector $\mathbf{\hat p_t}= (\hat x_t, \hat y_t)$ where $\hat x_t$ and $\hat y_t$ are the predicted probability of the click and purchase action on $I_t$.
The recurrent model is implemented by one stack layer LSTM \cite{sundermeyer2012lstm} with the hidden size of 256, and we unroll the LSTM cell in a maximum sequence length of 25.
We optimize the simulator network using the sum of cross-entropy loss between the ground-truth $\mathbf{p_t}$ and $\mathbf{\hat{p}_t}$ across all time-steps.

\section{Experiment}

We showcase the effectiveness of our approach in a series of simulated experiments and live experiments in a real-world Taobao ad system. We consider two scenarios in the homepage of Taobao App: 1) \textit{Good Items} targets the consumers with a high expense, so the ad items are usually in high quality; 2) \textit{Guess What You Like} aims to perform personalized advertising strategies, and thus the items are chosen based on users' preferences, interests, and recent behaviors.

\subsection{Empirical Evaluation: Simulations}

The dataset\footnote{Dataset is available: \color{blue}\url{https://github.com/465935564/sequential_advertising_data}} includes 58,648 request sessions from 4,988 sampled users in the two scenarios within three days. 
Each request contains a candidate ad set $\mathcal{D}$ ($50 \le |\mathcal{D}| \le 400$). 
The whole dataset involves 52,749 ad items and 4,543,880 records in total.

Each category has 5 ads on average.
As each scenario has only one ad position, we have $K$=1.
All the request sessions of a consumer are sorted in session time to form a consumer trajectory. 
We use 90\% of the trajectories as a training set while the rest 10\% leaves for test evaluation.

\subsubsection{Simulator Training}

\begin{figure}
\begin{subfigure}{.49\columnwidth}
\includegraphics[height=3cm]{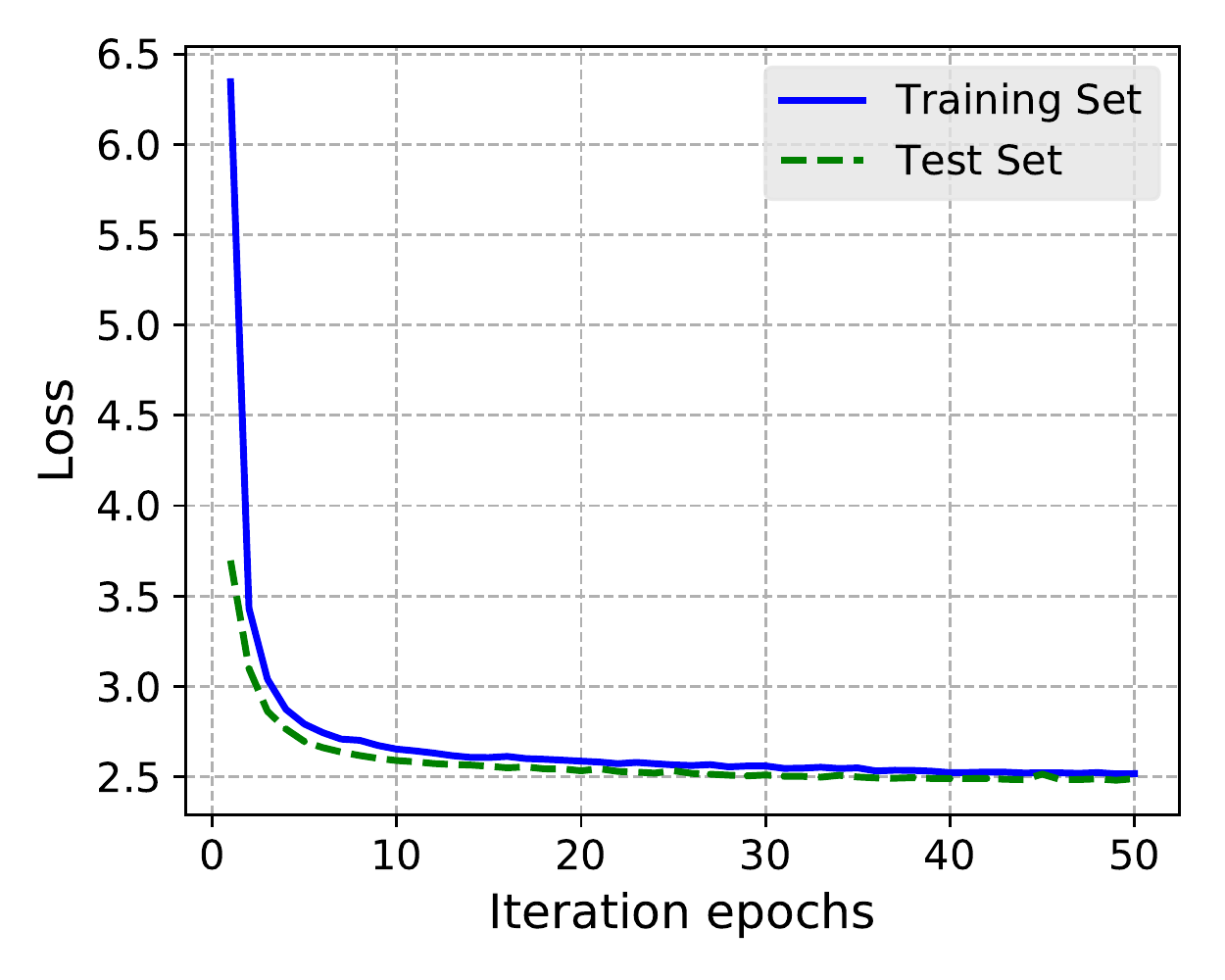}
\caption{Simulator Learning (loss).}
\end{subfigure}\hfill
\begin{subfigure}{.49\columnwidth}
\includegraphics[height=3cm]{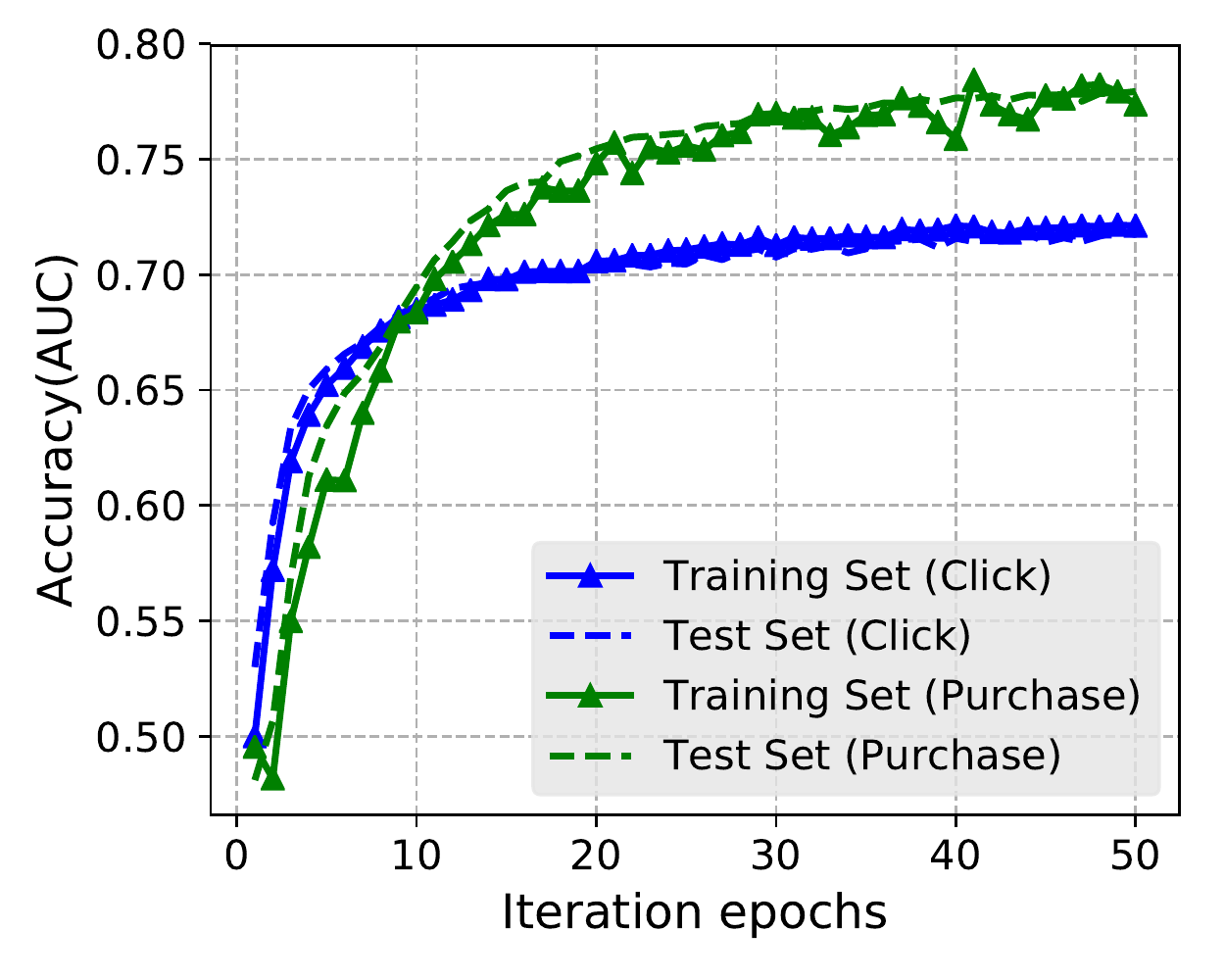}
\caption{Simulator Learning (AUC).}
\end{subfigure}
\begin{subfigure}{.49\columnwidth}
\includegraphics[height=3cm]{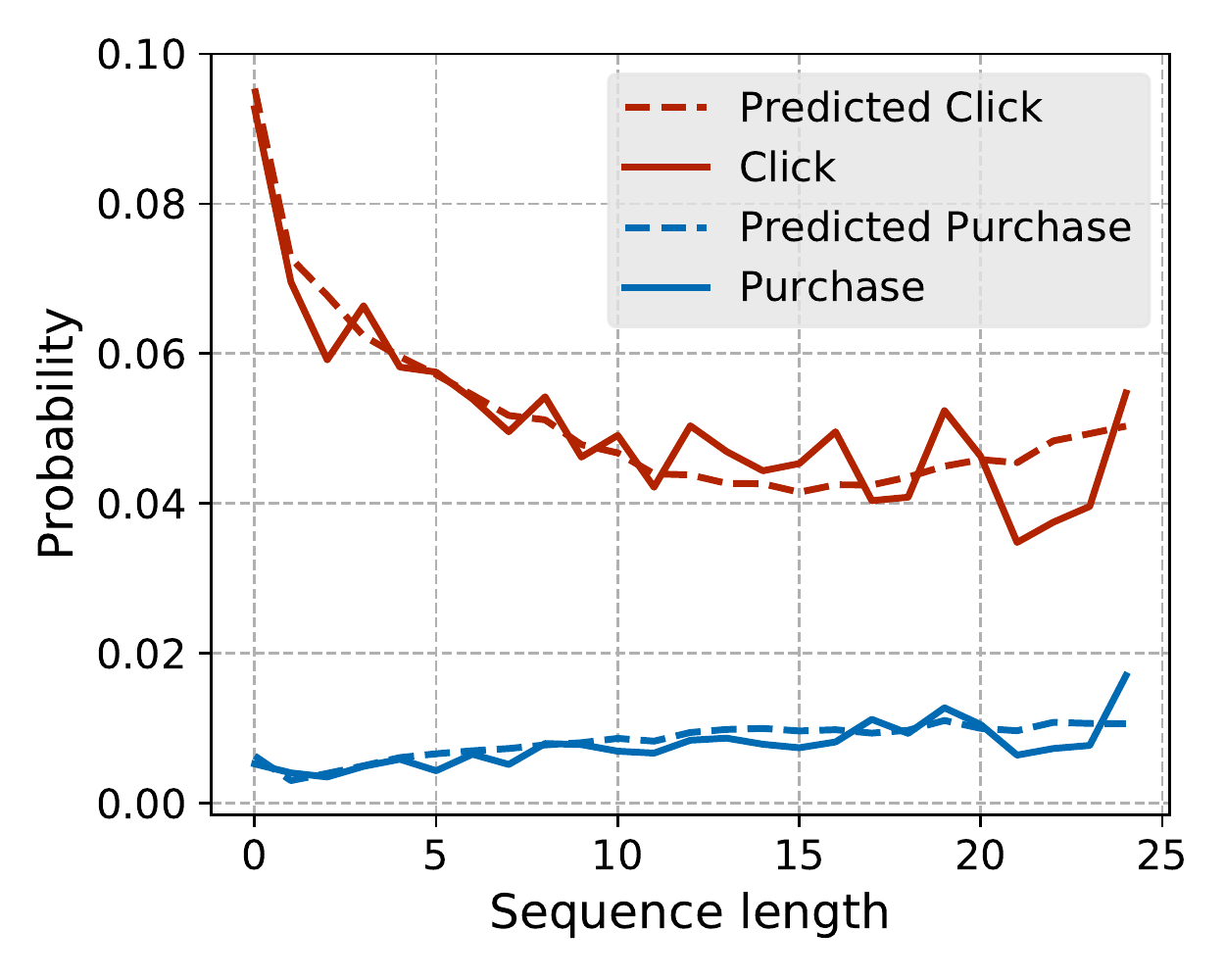}
\caption{Simulator Predictions.}
\end{subfigure}\hfill
\begin{subfigure}{.49\columnwidth}
\includegraphics[height=3cm]{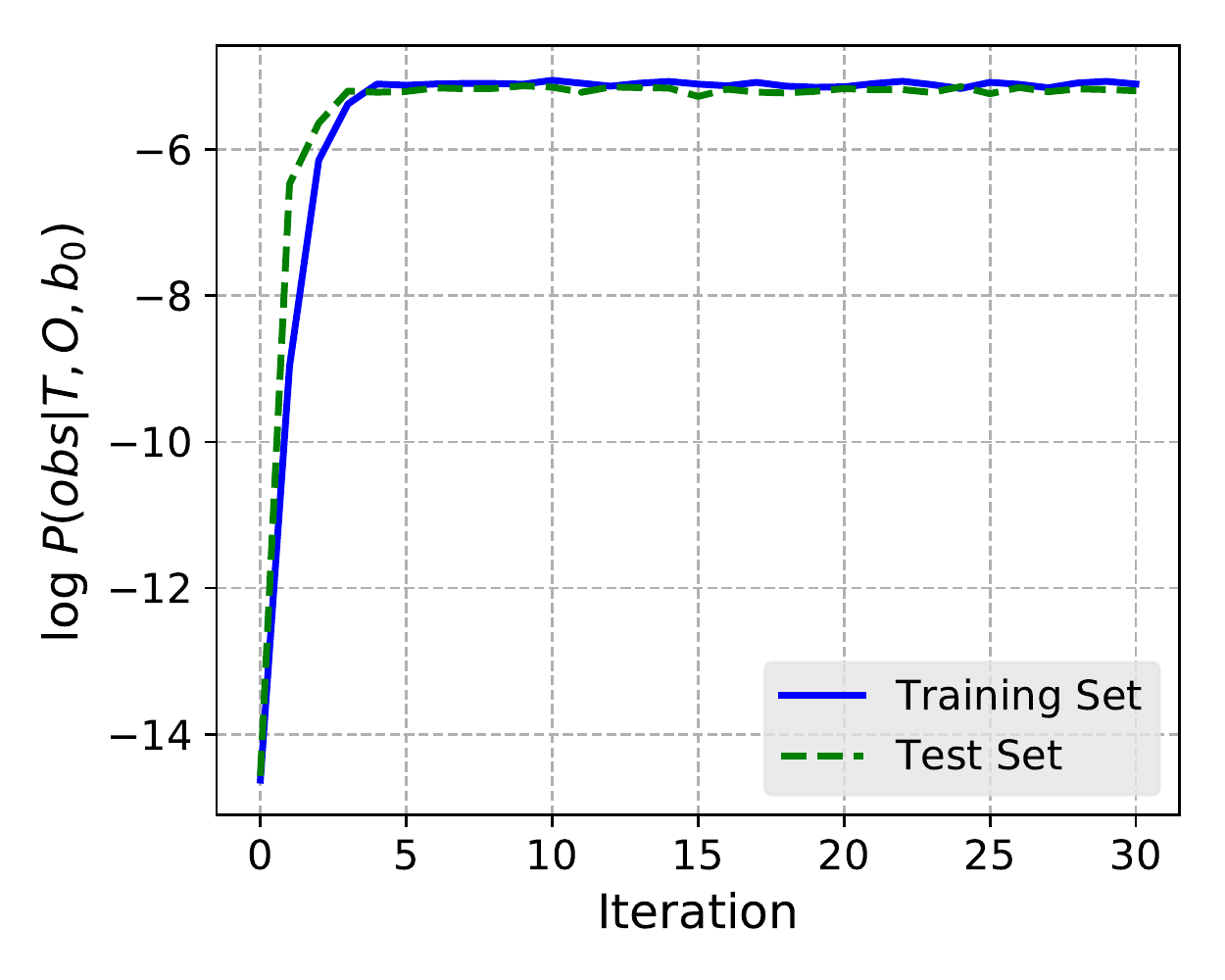}
\caption{Learning Curves of EM model.}
\end{subfigure}
\caption{Training result of the simulator and EM model.}
\label{fig:simulator}
\end{figure}

To conduct offline experiments, we train an environment simulator to imitate user click and purchase actions on an ad item.
When the agent decides on an item for a user, our simulator will generate the click and conversion rate for this ad-user pair, from which we sample the final click/purchase actions.
The consistency of simulated data and the real-world data is important, and thus we evaluate the simulator in 3 ways:

We first show the learning loss of the training and test set in Fig. \ref{fig:simulator}(a), which illustrates the loss converges well.
The learning accuracy (AUC score) of two predictions are given in Fig. \ref{fig:simulator}(b). 
We achieved 0.732 AUC for click and 0.771 AUC for purchase at 50-th epochs, which proves the prediction ability of our learned simulator.
Apart from the accuracy curves, we compare the simulated prediction with the ground-truth data, depicted as Fig. \ref{fig:simulator}(c).
The figure shows that the simulator can correctly predict the trends of real data.
Beyond that, we also find an interesting phenomenon: the conversion rate (the blue line in Fig. \ref{fig:simulator}(c)) will increase if we impress a user by repeated displays, which indicates the potential benefits of sequentially repeated advertising exposures. 

\subsubsection{Policy Learning and Evaluation}
In this part, the agent optimizes its advertising strategies based on the user latent states by the feedback provided by the simulator.
To infer user states, we train and evaluate the EM model by the log probability curves of the observed sequences in the test set shown in Fig. \ref{fig:simulator}(d), where the parameters converge well. 
The number of $|\mathcal{S}|$ controls how fine we split users' latent states $\mathcal{S}=\{s_1,s_2,...s_j\}$. 
In our case, we use $|\mathcal{S}| = 3$ because we find the converged log probability does not increase much when $|\mathcal{S}| > 3$. 
The following algorithms are compared with our method with the same observations, actions, and rewards settings\footnote{Due to our settings in discrete-actions and memory replays, we do not consider continuous-action or asynchronous-specific RL techniques, such as DDPG, A3C, etc.}.

\begin{table}
\footnotesize
\begin{tabular}{c |c |c c c c }
\hline
 Parameter & Method & Revenue & Cost & ROI & Reward \\
\hline
- &  Manual bid & 100\% & 100\% & 100\% & 100\% \\
\hline
- & Bandit & 107.6\% & 99.6\% & 108.1\% & 112.8\% \\
\hline
\multirow{4}{*}{$\gamma$=0.1} & 
     DQN  & 99.3\% & 99.1\% & 100.2\% & 101.4\%\\
 &  EM-DQN & 92.2\% & 91.1\% & 101.2\% & 103.3\% \\
 &  ADRQN & 95.5\% & 97.1\% & 98.3\% & 107.2\% \\
 &  DISA & \textbf{100.3\%} & \textbf{98.2\%} & \textbf{102.2 \%} & \textbf{107.9\%}\\
\hline
\multirow{4}{*}{$\gamma$=0.3} & 
   DQN & 104.5\% & 100.9\% & 103.5\% & 112.9\%\\
&   EM-DQN & 105.1\% & 101.2\% & 103.8\% &107.8\% \\
&  ADRQN & 110.7\% & 101.3\% & 109.2\% &114.5\% \\
 &  DISA & \textbf{111.7\%} & \textbf{100.7\%} & \textbf{110.9\%} & \textbf{115.9\%}\\
\hline
\multirow{4}{*}{$\gamma$=0.5} & 
   DQN & 110.3\% & 101.2\% & 109.0\% & 114.6\% \\
& EM-DQN & 111.2\% & 101.7\% & 109.2\% &114.5\% \\
&  ADRQN & 112.9\% & 102.7\% & 110.0\% &116.7\% \\
 &  DISA & \textbf{113.8\%} & \textbf{101.1\%} & \textbf{112.5\%} & \textbf{117.5\%} \\
\hline
\multirow{4}{*}{$\gamma$=0.7} & 
  DQN & 109.9\% & 101.5\% & 108.3\% & 113.3\%\\
 & EM-DQN & 109.9\% & 101.0\% & 108.8\% & 112.8\% \\
 & ADRQN & 116.8\% & 103.4\% & 112.8\% & 120.0\% \\
 & DISA  & \textbf{119.0\%} & \textbf{101.9\%} & \textbf{116.7\%} & \textbf{122.1\%} \\
\hline
\multirow{4}{*}{$\gamma$=0.9} & 
  DQN & 112.3\% & 101.9\% & 110.1\% & 115.7\% \\
&  EM-DQN & 113.4\% & 102.1\% & 111.0\% & 116.2\% \\
&  ADRQN & 117.0\% & 103.0\% & 113.6\% & 121.0\% \\
 &  DISA & \textbf{120.9\%} & \textbf{100.7\%} & \textbf{120.0\%} & \textbf{125.2\%}\\
\hline
\end{tabular}
\caption{Performance under different parameter settings.}
\label{table:performance}
\end{table}

\textbf{Manual bid}. It's the bid strategy using humans' experience \cite{jin2018real}. 

\textbf{Bandit}. Contextual bandit \cite{allesiardo2014neural} is an online algorithm that maximizes the total payoff of the chosen actions given the context.

\textbf{DQN}. DQN \cite{dqn} is a model-free RL algorithm. It directly takes in the observations and outputs the ranking policy by selecting the largest Q-value action.  

\textbf{ADRQN}. It is a recurrent variant of DQN where the current observation and the last time-step action are fed to an LSTM network \cite{zhu2018improving}. This is a model-free POMDP where the latent state is implicitly captured and modeled by the LSTM.

\textbf{DISA}. This is our proposed model-based POMDP algorithm. 
DISA explicitly estimates the beliefs (distribution of hidden states) and learns to optimize its policy by the belief value approximation.

\textbf{EM-DQN}. It is a variant of DQN where its input is the beliefs of DISA rather than observations. This method attempts to learn the mappings from beliefs to actions with the model-free RL.

\begin{figure}
\begin{subfigure}{.49\columnwidth}
\centering
\includegraphics[height=2.8cm]{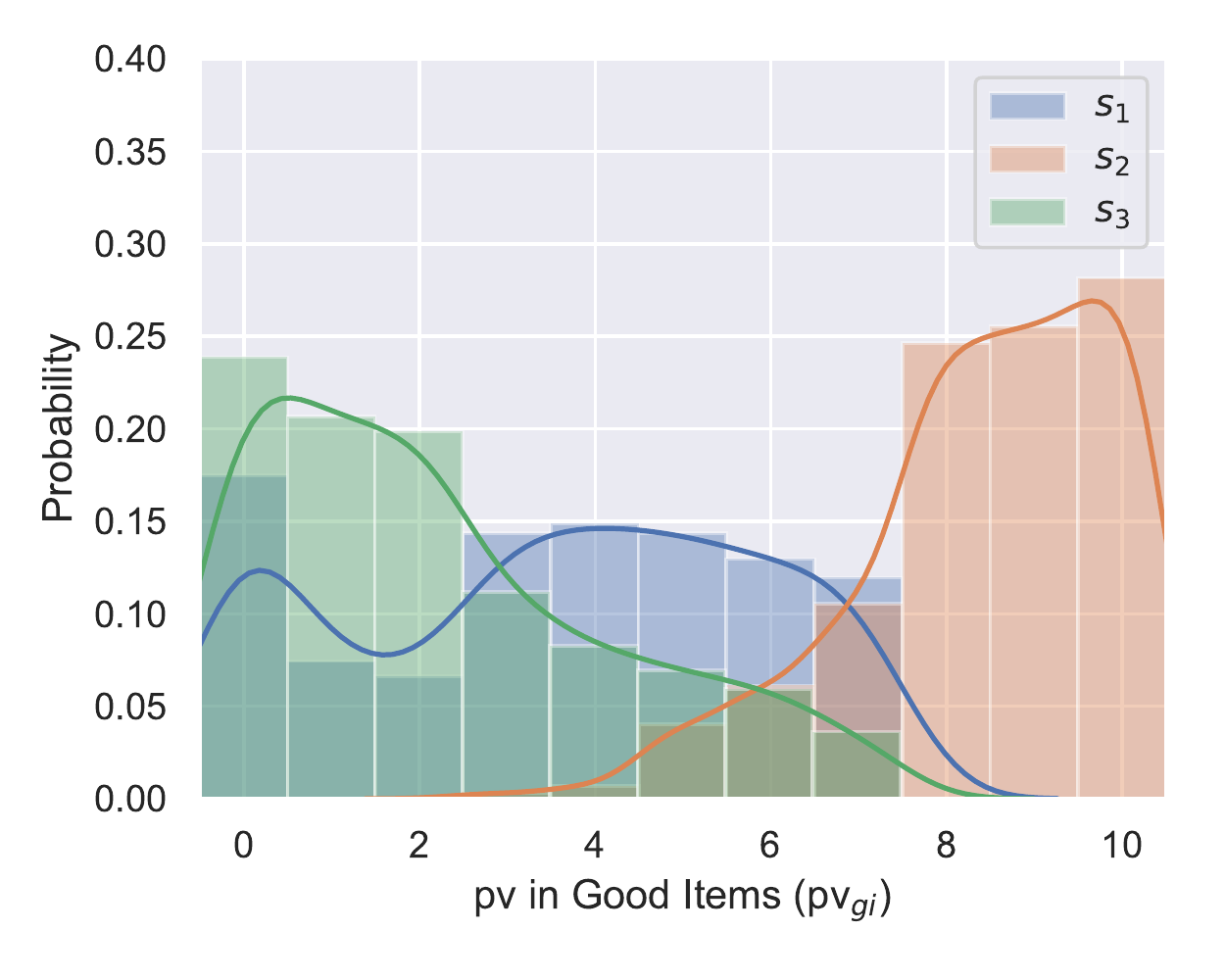}
\end{subfigure}
\begin{subfigure}{.49\columnwidth}
\includegraphics[height=2.4cm]{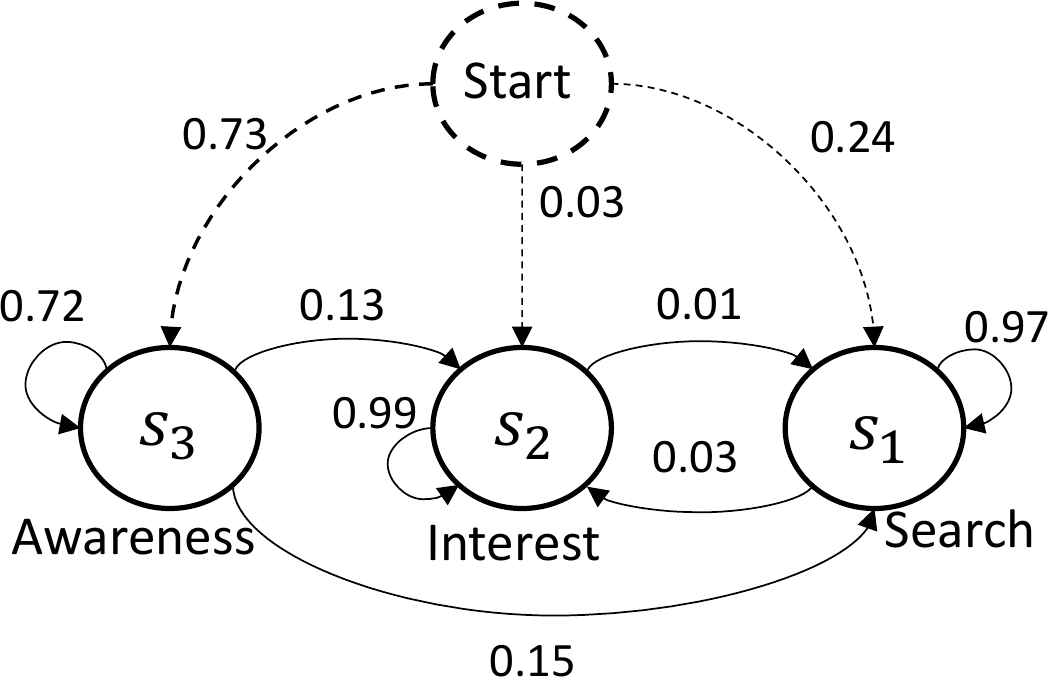}
\end{subfigure}
\caption{Distributions of the learned parameter $O(o_1|s_i)$ (Left), and the diagram for state transitions (Right).}
\label{fig:state_trans}
\end{figure}

\begin{table}
\begin{center}
\resizebox{\columnwidth}{!}{%
\begin{tabular}{c|c|c c c c c|c c c c}
\hline
\multirow{3}{*}{State} & Init & \multicolumn{5}{|c}{Observation  $\mathbb{E}[O(o_j|s_i)]$} & 
    \multicolumn{4}{|c}{Transition $T(s'|s)$} \\
\cline{2-10}
  & \multirow{2}{*}{$b_0$} 		& $o_1$				& $o_2$ 				& $o_3$				& $o_4$  				& $o_5$ 	 & \multirow{2}{*}{$s'_1$} &\multirow{2}{*}{$s'_2$} &\multirow{2}{*}{$s'_3$} \\
  &   						& (pv\textsubscript{gi}) 	& (clk\textsubscript{gi})	& (pv\textsubscript{gw})	& (clk\textsubscript{gw})	 &  (scen)  &           				&    				      & 			           \\
\hline
 $s_1$ & 0.24 		& 3.56 		& 0.19 		& \textbf{1.90 }		& \textbf{0.21} 		&  \textbf{0.78} 	& 0.97 & 0.03 & 0 \\
$s_2$ & 0.03 		& \textbf{8.41} 	& \textbf{0.71}   & 0.98  			& 0.07			& 0.38  		& 0.01 & 0.99 & 0  \\
$s_3$ &\textbf{0.73} & 2.22 			& 0.06  		& $\approx$ 0 		& $\approx$ 0 		& 0.02 		& 0.15 & 0.13 &  0.72  \\
\hline
\end{tabular}
}%
\footnotetext{Footnote}
       \begin{tablenotes}
       	\footnotesize
            \item Here, each $o_j$ refers to one dimension of $o$ vector, that is $O(o_j|s_i) = O(o[j]|s_i)$
        \end{tablenotes}
\end{center}
\caption{Statistics of the learned parameters in SE.}
\label{table:parameters}
\end{table}

For fair comparisons, several experiments are conducted to show the performance of different methods with the same $\gamma$ parameter in Table \ref{table:performance}. 
Each method is evaluated by the ROI indicator (revenue/cost) and the average rewards (advertisers' profits). 
A higher ROI shows the stronger ability of earning more income with the same budget cost.
A higher reward is also important as it indicates a method can help advertisers obtain more profits.

From Table \ref{table:performance}, almost all the RL-based methods achieve higher ROI than Bandit method with $\gamma \in \{0.5, 0.7, 0.9\}$ since their decision-making is based on the long-term rewards. 
Under the same setting of $\gamma$, DISA outperforms all the others in ROI while achieving almost the same cost as other baselines. 
These results indicate the superiority of DISA as it not only helps advertisers earn more income per budget cost but also improves profits. 
Compared with DQN, for all $\gamma$, a higher ROI of EM-DQN shows the benefits of inferring beliefs over the behavior-action mappings (black-box) in model-free fashion.
Furthermore, DISA also demonstrates its advantage of the belief value approximation in SPOVA over the general neural network (pure belief-action mappings) in EM-DQN by ROI.

\begin{figure*}
\begin{subfigure}{.27\textwidth}
\centering
\includegraphics[height=3.3cm]{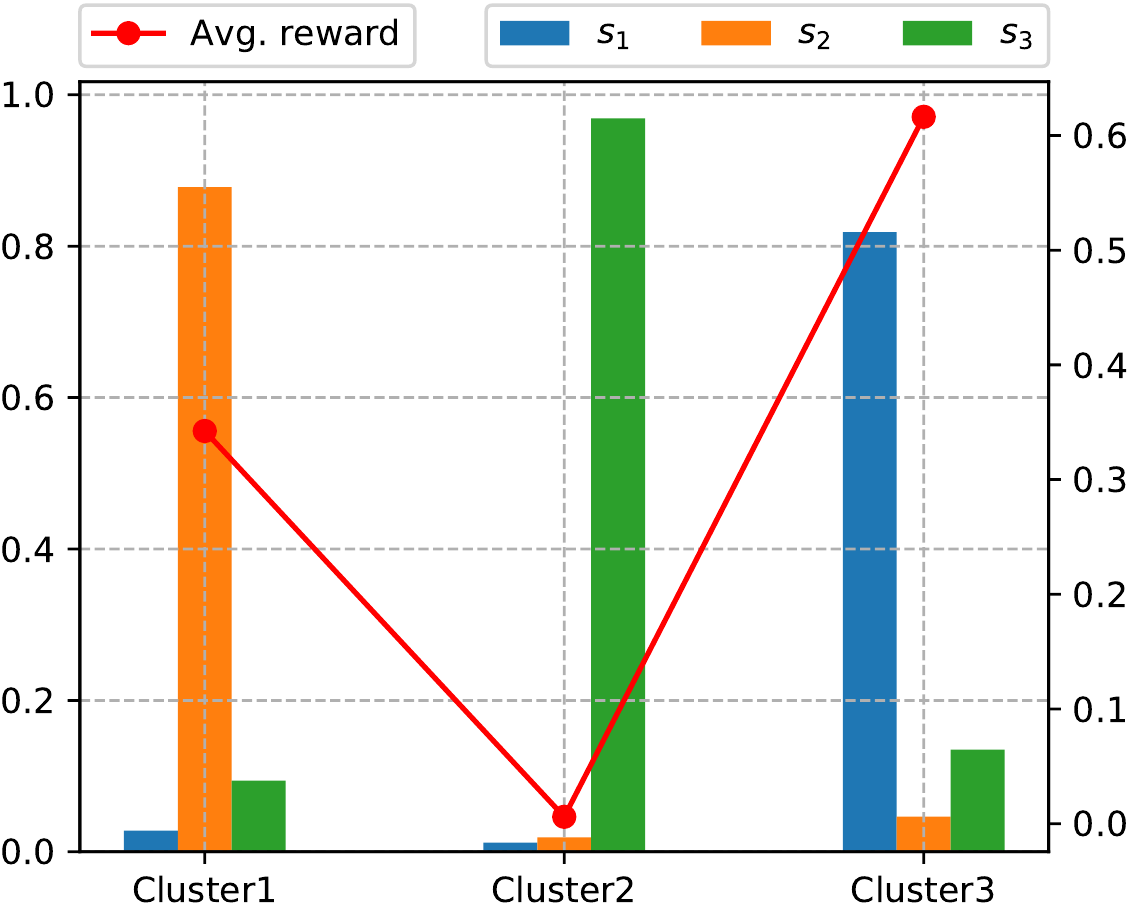}
\caption{Clustering statistics.}
\end{subfigure}\hfill
\begin{subfigure}{.40\textwidth}
\centering
\includegraphics[height=3.1cm]{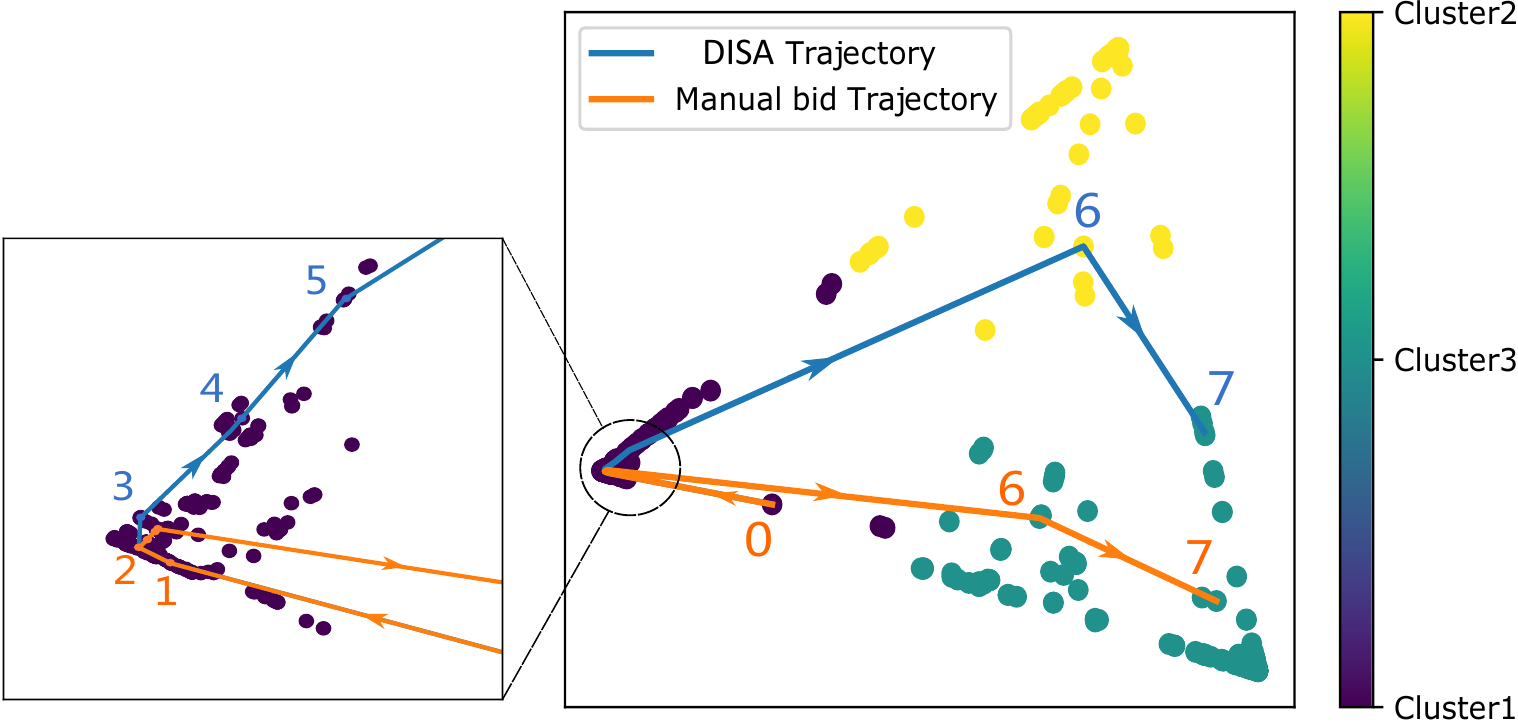}
\vskip 0.65em
\caption{Evolutionary trajectories of beliefs in 2-d projection.}
\end{subfigure}\hfill
\begin{subfigure}{.30\textwidth}
\centering
\includegraphics[height=3.45cm]{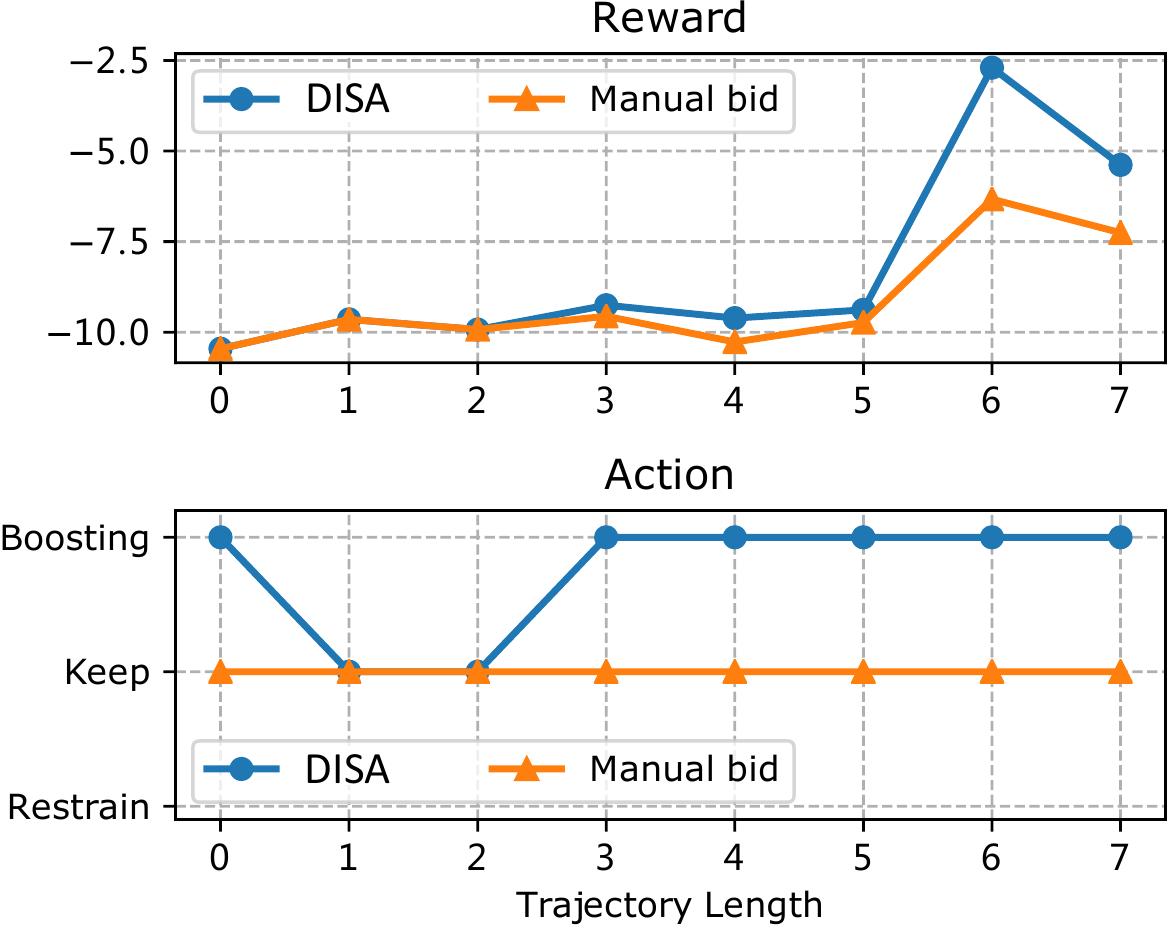}
\vskip -0.355em
\caption{Advertising actions and rewards.}
\end{subfigure}
\caption{Belief clustering and the evolutionary trajectories with different strategies.}
\label{fig:interpretation}
\end{figure*}

\subsubsection{Interpretations of Learned Hidden States}

Essentially, the EM learns a mapping from high-dimensional historical observations/actions to a compressed belief state, and this mapping is reflected in the learned parameters $T(s'|s_i,a)$, $O(o|s_i,a)$ and $b_0(s_i)$.
By analyzing these parameters, we can know how each state connects with different observations, so we can further interpret the property of each state $s_i$. 
To do this, one direct way is to compare the distribution\footnote{For better explanation, we slightly abuse the notation in this section. We marginalize out $O(o|s,a)$ and $T(s'|s,a)$ for all $a$ to obtain $O(o|s)$ and $T(s'|s)$.}of an observation $O(o|s)$ w.r.t each state, e.g., Fig. \ref{fig:state_trans}(Left) illustrates that a large value of $o_1$ is more likely to be observed under $s_2$ rather than $s_3$ and $s_1$, so we can distinguish $s_2$ by the large value of the expectation\footnote{We define $\mathbb{E}[O(o|s,a)] = \sum_i O(o_i|s,a)o_i$ where $o_i$ is the observed value.} $\mathbb{E}[O(o_1|s_2)]$.
According to such different expectations of each observation, we can easily explain the characteristics of each state.

In our ad system, the observations reflecting a user's intent mainly include the number of exposure, click and purchase of the ad to the user, as well as how the user behaves in different scenarios. More concretely, our observations are that: \textit{pv\textsubscript{gi}} and \textit{clk\textsubscript{gi}} represent how many previous exposure and clicks of an ad have been made in \textit{Good Items} (similar for \textit{pv\textsubscript{gw}} and \textit{clk\textsubscript{gw}} in \textit{Guess What You Like}), and $scen$ describes how frequently a user switches to other scenarios. Here, we neglect purchase observations as the data is too sparse. Table \ref{table:parameters} lists the learned parameters in Section 6.1.2 w.r.t these observations, so now we can interpret each state as following:

State $s_3$ is an \textbf{awareness state} since the users under $s_3$ are observed to have little advertising exposure and clicks, particularly in \textit{Guess What You Like}. ($\mathbb{E}[O(o_3|s_3)] \approx 0$, $\mathbb{E}[O(o_4|s_3)] \approx 0$).
State $s_2$ is an \textbf{interest state} because we observe a large number of user browsing and click behaviors in this state, especially in \textit{Good Items} ($\mathbb{E}[O(o_1|s_2)] = 8.41$, $\mathbb{E}[O(o_2|s_2)] = 0.71$).
Compared with state $s_2$, state $s_1$ is more active because the users are more likely to switch to \textit{Guess What You Like} while maintaining a relative high level of browsing behaviors  ($\mathbb{E}[O(o_5|s_1)]=0.78$, $\mathbb{E}[O(o_3|s_1)]=1.90$, $\mathbb{E}[O(o_1|s_1)] > \mathbb{E}[O(o_1|s_3)]$); 
this explains that users in $s_1$ start to actively search for their interested items across different scenarios, and thus we label $s_1$ as a \textbf{search state}. 
Note that our analysis is compatible with the definition of \textit{customer funnel} revealed in \cite{noble2010s,abhishek2012media,jansen2011bidding,ghose2015towards}, and the differences are that: 1) our results are data-driven and learned from a validated EM model, and 2) we treat the final conversion state as an observable state instead of a latent state that requires inference.

Furthermore, we can also verify our interpretations above by $b_0$ and $T(s'|s)$, depicted in Fig. \ref{fig:state_trans}(Right).
$b_0$ tells us that almost 73\% of users start from the awareness state, while 24\% of users begin with the search state. 
$T(s'|s)$ describes how each state transits: i) 
awareness $s_3$ $ \yrightarrow{15\%}[-2pt] $ search $s_1$ 
$ \yrightarrow{ 3\%}[-2pt] $ interest $s_2$,
and  ii) awareness $s_3$ $ \yrightarrow{13\%}[-2pt] $ interest $s_2$. 
These transition routes indicate that a user's status always transits from awareness to interest/search rather than going in reverse, which is consistent with our common sense.

\subsubsection{Interpretations of Learned Strategies}
Based on the interpretable state, we can compare the difference of the belief's evolutionary tracks by performing two different advertising strategies (DISA, Manual bid) on the same user trajectories.

We collect all the inferred belief vectors and project them into a 2-dim space with PCA techniques as Fig. \ref{fig:interpretation}(b).
For better visualization, we use K-means to cluster those nodes into 3 clusters with different colors so that each cluster is dominated by one type of hidden state, e.g., more than 90\% of the belief nodes in cluster1 belong to state $s_3$, depicted as Fig. \ref{fig:interpretation}(a).
So we can label each cluster with the property of each state: cluster 1, cluster 2 and cluster 3 are regarded as an \textbf{awareness stage} ($s_3$), an \textbf{search stage} ($s_1$) and an \textbf{interest stage} ($s_2$) respectively. 
Furthermore, we compare the average reward collected at each stage in Fig. \ref{fig:interpretation}(c), which shows the search/interest stage earns much higher rewards than the awareness stage; this in turn proves the rationality of our analysis on each state. 

Let's examine a typical trajectory where consumers browse dress items in \textit{Good Items} first with 6 requests and then in \textit{Guess What You Like} with 2 more requests. 
Fig. \ref{fig:interpretation}(b) gives two evolutionary trajectories of their states under the strategy of DISA and the manual bid baseline. 
We can see that both two trajectories start from the awareness stage and also end in the interest stage, but they get separated after the 3-rd advertising action. 
This separation leads to the main difference of two trajectories: DISA successfully guides the hidden state transiting to the search stage while the human bid baseline does not. 
We draw the performed actions and corresponding rewards in Fig. \ref{fig:interpretation}(c), which shows that the boosting actions in DISA dominate after the 3-rd action. 
One reasonable explanation is that: the boosting action can guarantee the display of ad items and further impact the consumer's perception on the items, especially in \textit{Good Items}. 
Therefore, after the consumer switches to \textit{Guess What You Like}, the repeated boosting on the same item helps transit consumer's state to the search stage, which leads to a relatively higher reward as shown in Fig. \ref{fig:interpretation}(c).

\begin{figure}
\centering
\begin{subfigure}{.49\columnwidth}
\includegraphics[height=2.5cm]{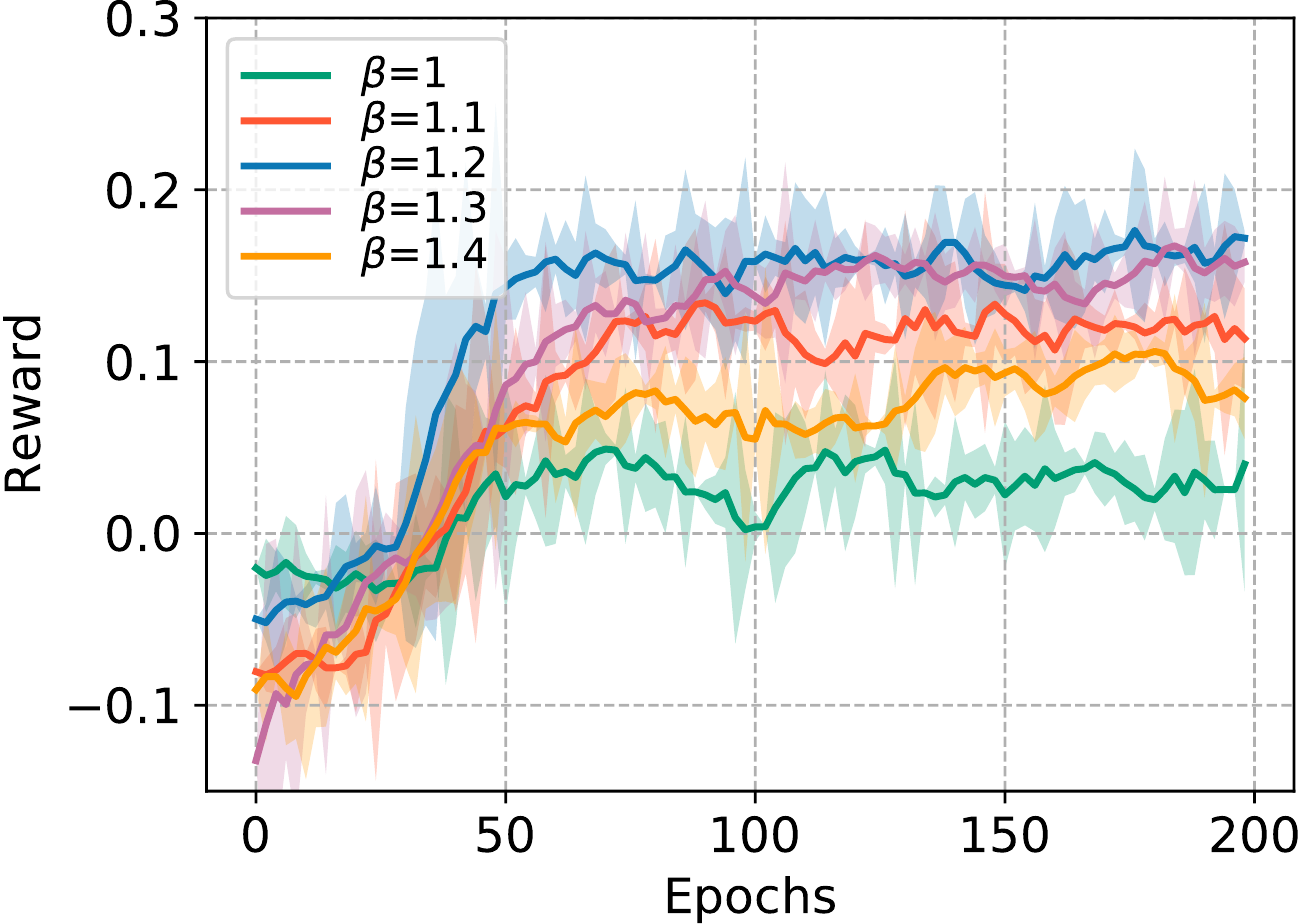}
\end{subfigure}
\begin{subfigure}{.49\columnwidth}
\includegraphics[height=2.5cm]{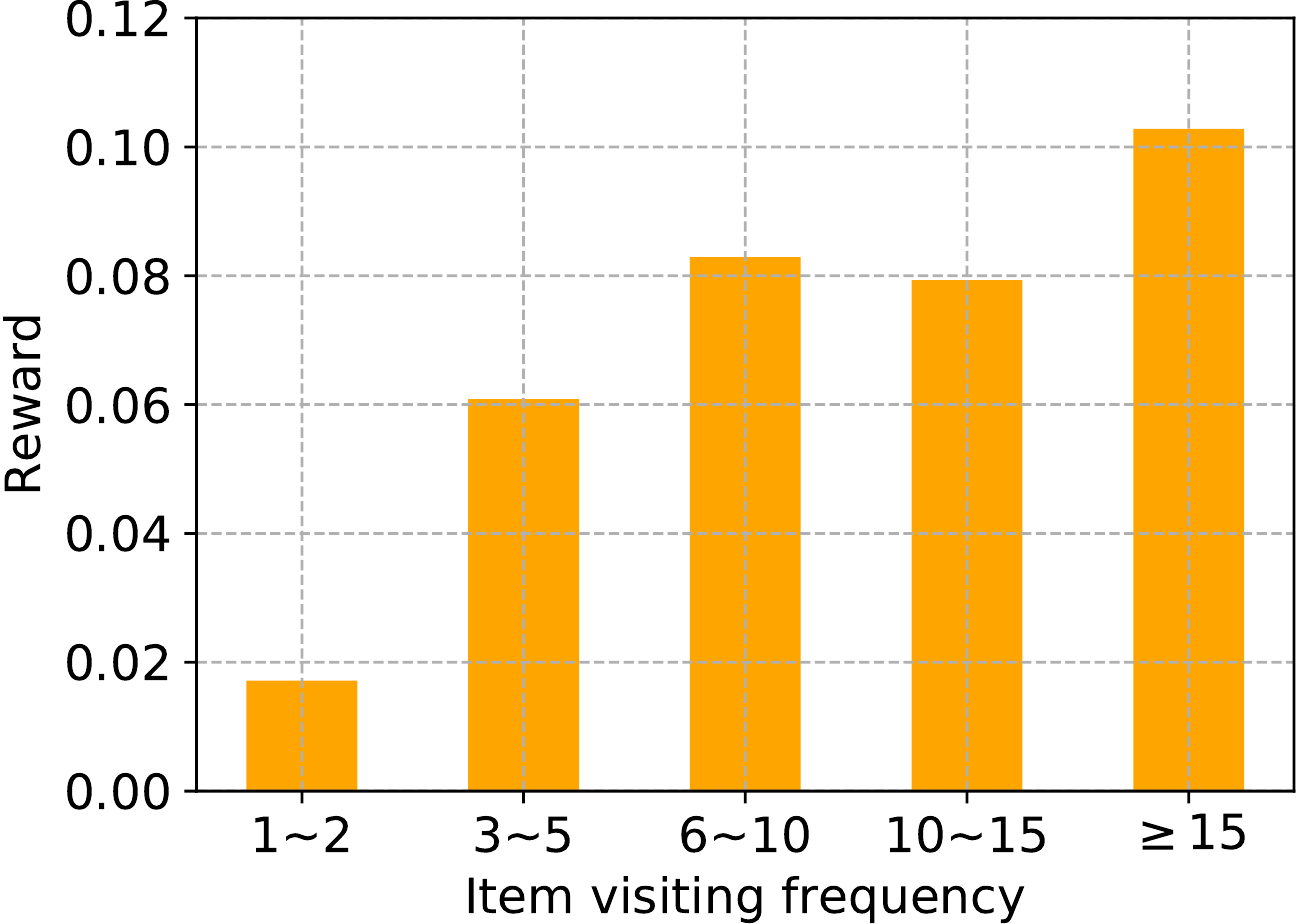}
\end{subfigure}
\caption{Rewards under different $\beta$ (Left), and rewards under different visiting frequencies (Right).}
\label{fig:beta}
\end{figure}

\subsubsection{Reward Settings} 
The value of $\beta$ determines the degree of punishment for performing the boosting action ($\beta=1$ means no punishment).
With small $\beta$, the agent is easier to use boosting action to win the bidding, leading to the increase of impressions/cost and further reaching low rewards. 
Large $\beta$ means fewer impression opportunities to obtain revenue and will also achieve low rewards. 
In Fig. \ref{fig:beta} (Left), we find $\beta=1.2$ can well control the frequency of boosting actions so that rewards are maximized. 
In our data, click behavior happens 5-10 times more than purchase (shown in \ref{fig:simulator}(c)), and therefore, $\lambda=5$ is enough to adjust the data magnitude between revenue and cost; 
besides, we also find $\lambda=5$ performs best in ROI by the parameter grid search.  
The window $T_w$ is set to 3 hours since we find 90\% conversions are reached within 3 hours.

\subsubsection{Performance within Different Items} We compare the rewards under the items with different visiting frequencies in Fig. \ref{fig:beta} (Right). It is clear that the more a user interacts with an item, the more reward is gained. However, when the visiting frequency is less than 2, the reward becomes much lower, which can be reasoned that it is hard to transfer users to the interest/search state with only two steps. It also shows our model works better with a longer sequence.

\begin{figure}
\centering
\includegraphics[height=2.3cm]{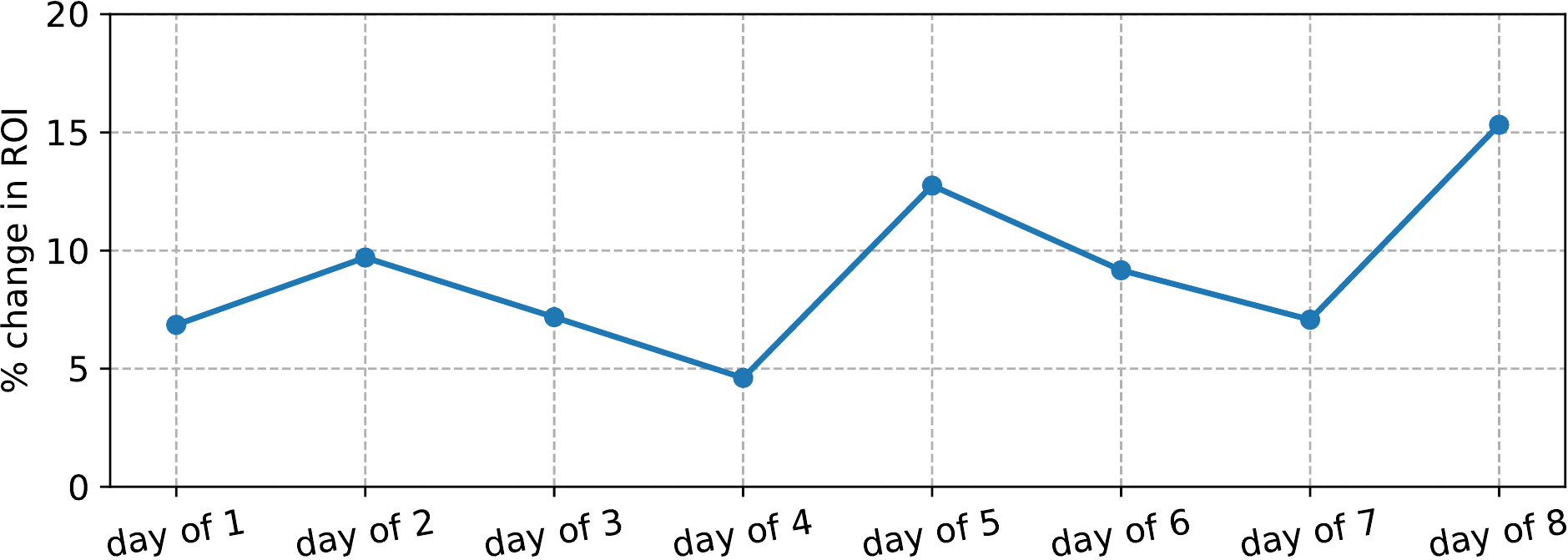}
\caption{Increase in ROI over the control group.}
\label{fig:online}
\end{figure}

\subsection{Live Experiments}

We conduct online A/B experiments running in the live ad platform. 
Experiments are run from Oct.26 to Nov. 2 in 2019, which involves randomly sampled 9,165,752 users, 664 advertisers, and 72,381 ad items from 12,401 categories.
Our sequential advertising model (experimental group) is trained continuously using all user behaviors across 9 scenarios with a lag under 24 hours. 
The control group is a deployed production model (Cross Entropy Method, CEM \cite{de2005tutorial}) that optimizes for immediate rewards. 
We allocate the same budget cost to the control and experimental group for each advertiser (we asked the advertisers for permission to adjust their budgets).
We focus our discussion on the amount of revenue and ROI of the advertisers.
In Fig. \ref{fig:online}, we achieved +9.02\% of revenues with the same budget cost (-0.81\%), resulting in +9.75\% of ROI for the experimental group.
As our live results are promising, our algorithm has been officially deployed online and allows advertisers to customize their advertising strategies.

\section{Conclusions}
In this paper, we proposed our DISA to model the sequential advertising problem, which optimized the strategies by taking account of  interpretability. 
We developed POMDP framework in large-scale industrial settings to infer hidden states based on the consumer's historical behaviors. 
To best fit our interpretable model, a variant of SPOVA based on deep neural networks has been proposed to learn value function and optimize advertising policies. 
Many details of our implementation were provided. 
The simulation and A/B online results have validated the superiority of the proposed algorithm against several DRL baselines. 
In several cases' analysis, we try to interpret the learned hidden states, which are meaningful and consistent with our business common sense.

\bibliographystyle{ACM-Reference-Format}
\bibliography{sample-base}

\appendix
\section{Supplementary}
\subsection{Derivation of parameter learning for DISA}
Let us consider a discrete extended HMM in session 5.1 with length $L$. Let the space of observations, hidden states, and actions  be $M$, $N$, and $A$ respectively. Given a sequence of observations $\mathcal{O}=\{o_1 \sim o_{L}\}$ and corresponding actions $\mathcal{A}=\{a_0 \sim a_{L-1}\}$, a POMDP model is parameterized by a extended HMMs with $\theta=(b_0, T, O)$. Specifically, $b_0(i)=P(s_1=i)$ is the initial state distribution, $T_{i, k} ^ {j} = P(s_{t+1} = j, s_{t}=i | a_t= k)$ is the transition function, and $O_{i, k}(j)= P(o_{t} = j, s_{t}=i | a_{t-1}= k)$ is the observation function. The Q-function is defined as the expectation term that we need to maximize:
\begin{equation} \notag
Q(\theta, \theta ^t ) = q(s)\sum_{s \in \mathcal{S}} \log P(\mathcal{O},s|\mathcal{A}; \theta^t)= E_{q(s)} \log P(\mathcal{O},s|\mathcal{A};  \theta ^t )
\end{equation}

\subsubsection{Extension of Baum-Welch procedures}
We extend Baum-Welch procedure for estimating $\theta^*$ from $\mathcal{O}$ and $\mathcal{A}$. Our method can be described as repeating the following steps until convergence:

\begin{enumerate}
\item E-step: compute $Q(\theta, \theta^{t})= \sum_s \log [ P(\mathcal{O}, s | \mathcal{A}; \theta)] P(s|\mathcal{O}, \mathcal{A} ; \theta^{t})$
\item M-step: set $\theta^{t+1} = \argmax_{\theta} Q(\theta, \theta^{t-1}) $
\end{enumerate}

Firstly, noting that $P(s, \mathcal{O}| \mathcal{A}) = P(s | \mathcal{O}, \mathcal{A}) P(\mathcal{O} | \mathcal{A})$, we can write the Q function as $\hat Q(\theta, \theta^{t})= \sum_s \log [ P(\mathcal{O}, s | \mathcal{A}; \theta) ] P(s,\mathcal{O} | \mathcal{A} ; \theta^{t})$ since $P(\mathcal{O} | \mathcal{A})$ does not affect the maximization of $Q$ in M-step. Now the $P(\mathcal{O}, s | \mathcal{A}; \theta)$ is easy to write: 

\begin{equation} \notag
\begin{aligned}
P(\mathcal{O},\mathcal{S}|\mathcal{A} ;\theta) &= P(o_1 \sim  o_L, s_1 \sim s_L | a_0 \sim a_{L-1} ; \theta) \\
&= b_0(s_1) \prod_{t=2}^L T_{s_{t-1}, a_{t-1}} ^{s_t} \prod_{t=1}^L O_{s_t,a_{t-1}}(o_t) \\
\end{aligned}
\end{equation} 
Taking the log gives us: 
$$
\log P(\mathcal{O},\mathcal{S}|\mathcal{A};\theta)= \log b_0(s_1) + \sum_{t=2}^L  \log T_{s_{t-1}, a_{t-1}} ^ {s_t} + \sum_{t=1}^L \log O_{s_t,a_{t-1}}(o_t)
$$
Plugging this into $\hat Q(\theta,\theta^{t})$, we get
\begin{equation} 
\begin{aligned}
\hat Q(\theta,\theta^{t}) &= \sum_s \log b_0(s_1) P(s,\mathcal{O}| \mathcal{A}; \theta^t)  \\
 &\phantom{{}=1} + \sum_s \sum_{t=2}^L \log T_{s_{t-1}, a_{t-1}} ^ {s_t} P(s,\mathcal{O}| \mathcal{A}; \theta^t) \\
 &\phantom{{}=1} + \sum_s \sum_{t=1}^L \log O_{s_t,a_{t-1}}(o_t) P(s,\mathcal{O}| \mathcal{A}; \theta^t) \notag\\
\end{aligned}
\end{equation}
Note that parameters are subjective to the constraints:
\begin{equation} 
\begin{aligned}
\notag
\sum_{s'} T_{s, a} ^ {s'} = 1 ; 
\sum_{o} O_{s, a} ^ {o} = 1 ;
\sum_{s} b_0(s) = 1   ;
\end{aligned}
\end{equation}
Applying Lagrange multiplier method, let $\hat L(\theta, \theta^t)$ be the Lagrangian 
\begin{equation} 
\begin{aligned}
\hat L(\theta, \theta^t) &= \hat Q(\theta, \theta^t) - \lambda_{b_0} \left ( \sum_{i=1} ^ N b_0(i) -1 \right ) - \sum_{i,k=1}^{N,A} \lambda_{T_{i,k}} \left ( \sum_{j=1} ^ N T_{i,k}^{j} -1 \right )  - \\
&\phantom{{}=1} \sum_{i,k=1}^{N,A}  \lambda_{O_{i,k}} \left ( \sum_{j=1} ^ M O_{i,k}(j) -1 \right ) \notag\\
\end{aligned}
\end{equation}
First let us focus on the $b_0(i)$. Let $\partial \hat L(\theta, \theta^t) / \partial b_0(i)=0 $ and  $\partial \hat L(\theta, \theta^t) \\ /  \partial \lambda_{b_0} =0 $, we obtain:
\begin{equation} 
\begin{aligned}
\notag
b_0(i) & = P(s_1= i |\mathcal{O} , a_0 ; \theta^t)
\end{aligned}
\end{equation}
Following a similar process for the $b_0$, we have:
\begin{equation} 
\begin{aligned}
\notag
T_{i,k}^j  &=\frac{\sum_{t=2}^L P(s_{t-1}= i, s_t= j | \mathcal{O} , a_{t-1}= k ; \theta^t)}{ \sum_{t=2}^L P(s_{t-1} = i |\mathcal{O} , a_{t-1}=k ; \theta^t)}
\end{aligned}
\end{equation}
The final thing is $O_{i,k}(j)$, which is slightly trickier, let $I(x)$ denotes an indicator function which is 1 if $x$ is true, 0 otherwise. Similar with $T_{i,k}^j$, we finally get:
\begin{equation} 
\begin{aligned}
\notag
O_{i,k}(j) &=\frac{\sum_{t=1}^L P(s_{t}= i |\mathcal{O} ,a_{t-1}= k ; \theta^t)I(x_t = j)}{ \sum_{t=1}^L P(s_{t} = i |\mathcal{O} , a_{t-1}=k ; \theta^t)}
\end{aligned}
\end{equation}
For brevity, we use simple denotations $\gamma(i,j,k)= P(s_{t-1}= i, s_t= j | \mathcal{O}, a_{t-1}= k ; \theta^t)$ and $\gamma(i,k)= \sum_{j=1}^N \gamma(i, j, k) = P(s_{t-1}= i | \mathcal{O}, a_{t-1}= k ; \theta^t)$. Note that $\gamma(i,j,k)$ and $\gamma(i,k)$ are both quantities and can be computed efficiently by a variant of forward-backwards algorithm for extended HMMs.

\subsubsection{Inference of extended HMMs} In order to compute the $\gamma(i, j, k)$, we need to solve the forward-backward pass, and the $\gamma$ algorithm in extended HMMs.

\textbf{Forward pass}: We use notations $\alpha(s_t,a_{t-1})$ ($t<L$) to represent the probability of being in hidden state $s_t$ given observations $o_1 \sim o_t $ and conditioned on $a_0 \sim a_{t-1}$, 
\begin{equation} 
\begin{aligned}
\notag
\alpha(s_t, a_{t-1}) &=  P(o_1 \sim o_t, s_t | a_0\sim a_{t-1}) \\
&= \sum_{s_{t-1}} \alpha(s_{t-1}, a_{t-2}) T_{s_{t-1}, a_{t-1}} ^ {s_t}  O_{s_t, a_{t-1}}(o_t)
\end{aligned}
\end{equation}
where $\alpha(s_1,a_0) = b_0(s_1) O_{s_1, a_0}(o_1)$  

\textbf{Backward pass}: Similarly, we use notations $\beta(s_t)$ ($t<L$) to represent the probability of observing $o_{t+1} \sim o_L $ conditioned on $s_t$ and $a_t \sim a_{L-1}$, 
\begin{equation} 
\begin{aligned}
\notag
\beta(s_t,a_t) &=  P(o_{t+1} \sim o_T | s_t, a_t\sim a_{L-1}) \\
&= \sum_{s_{t+1}} \beta(s_{t+1} , a_{t+1})  T_{s_{t},a_t}^{s_{t+1}} O_{s_{t+1}, a_t}(o_{t+1}) 
\end{aligned}
\end{equation}
where $\beta(s_{L-1}, a_{L-1})  = \sum_{s_L} T_{s_{L-1},a_{L-1}}^{s_L} O_{s_L, a_{L-1}}(o_L) $ 

\textbf{$\gamma$ algorithm}: after we recursively compute $\alpha(s_t ,a_{t-1})$ and $\beta(s_t,a_{t})$ for each $s_t$, we can easily obtain a $\gamma'(s_t,s_{t+1}, a_t)$ which is used to compute $\gamma(s_t,s_{t+1}, a_t)$,
\begin{equation} \notag
\begin{aligned}
\gamma'(s_t,s_{t+1}, a_t) & = P(s_t, s_{t+1}, o_1 \sim o_L | a_0 \sim a_{L-1}) \\
& = \alpha (s_t, a_{t-1}) T_{s_t,  a_t}^{s_{t+1}}  O_{s_{t+1}, a_t }(o_{t+1}) \beta(s_{t+1}, a_{t+1})
\end{aligned}
\end{equation}
Finally, we have: 
\begin{equation} 
\begin{aligned}
\notag
\gamma(i, j, k)
&=P(s_{t-1}= i, s_t= j | \mathcal{O}, a_{t-1}= k ; \theta^t) \\
&= \frac{\gamma'(s_{t-1}=i, s_t= j, a_{t-1} = k)}{ \sum_{i=1}^N \sum_{j=1}^N \gamma'(s_{t-1}=i, s_t= j, a_{t-1} = k)}
\end{aligned}
\end{equation}

\subsection{Experiment Details}

\begin{algorithm}[ht]
\caption{DISA}\label{alg:DISA}
Init a Q-network $Q_a(b ; \eta_a)$ for each action $a$ and a trajectory replay memory $D$\;
Init the estimator state with parameters $ \theta_0= (T, O, b_0)$\;
\For{e = 1 to E}{
Sample $M$ trajectories $ \mathcal{J} = \{J_1 \sim J_m \} $ from D \;
Construct $\mathcal{O}=\{o_1 \sim o_T\}$, $\mathcal{A}=\{a_0 \sim o_{T-1}\}$ from $\mathcal{J}$\;
\For{i=1 to $I$}{
$\theta^i = \argmax_{\theta} \mathbb{E}_{p(s|\mathcal{O},\mathcal{A};\theta^{i-1})} \log p(s,\mathcal{O}|\mathcal{A};\theta^{i-1}) $ 
until $\log p(s,\mathcal{O}|\mathcal{A};\theta^{i-1})$ does not increase \;
}
Update the state estimator $\theta_e = 0.99 \times \theta_{e-1}   +  0.01 \times \theta^I $  \;
Create a new user trajectory $J$ for a category\;
\For{t=1 to T}{
Get the current observation $o_t$ and previous $b_{t-1}$, $a_{t-1}$\;
Perform belief updating $b_t=SE(b_{t-1}, a_{t-1}, o_t ; \theta_e)$\;
With probability $\epsilon$ select a random action $a_t$ otherwise select $a_t = \argmax_a Q(b_t; \eta_a)$\; 
Execute $a_t$ and receive an reward $r_t$, and store a transition $\langle b_{t}, o_t, a_t, r_t\rangle$ into the trajectory $J$\;
Sample a minibatch of $N$ transitions $\langle b_j, a_j, r_j, b_{j+1}\rangle$\ from all trajectories in $D$\;
Update $\eta_a$ by minimizing loss with Eq. (9)
}
Update the trajectory replay memory $D= \{D \cup J \}$
}
\end{algorithm}

\subsubsection{Observation and Action Settings} 
To know the accumulated effect of a user repeated action in different scenarios, we also have a few features on top of the basic features.
In specific, for each user trajectory, we compute the accumulated \textit{pv}, \textit{pCVR} and \textit{click} to represent how many previous impressions and clicks have been made to a user for a category in different scenarios.
Let \textit{gw} denotes the the subscript of all accumulated features in \textit{Guess What You Like} while \textit{gi} denotes that in \textit{Good Items}, thus we have 6 more observation features:  \textit{pv\textsubscript{gw}}, \textit{pCVR\textsubscript{gw}}, \textit{clk\textsubscript{gw}}, \textit{pv\textsubscript{gi}}, \textit{pCVR\textsubscript{gi}}, \textit{clk\textsubscript{gi}}. 
In total, we use a 31-dim vector to describe a data record, which includes item-related features, session-related features, and accumulated features.

\begin{figure}
\centering
\includegraphics[page=4, height=2.5cm]{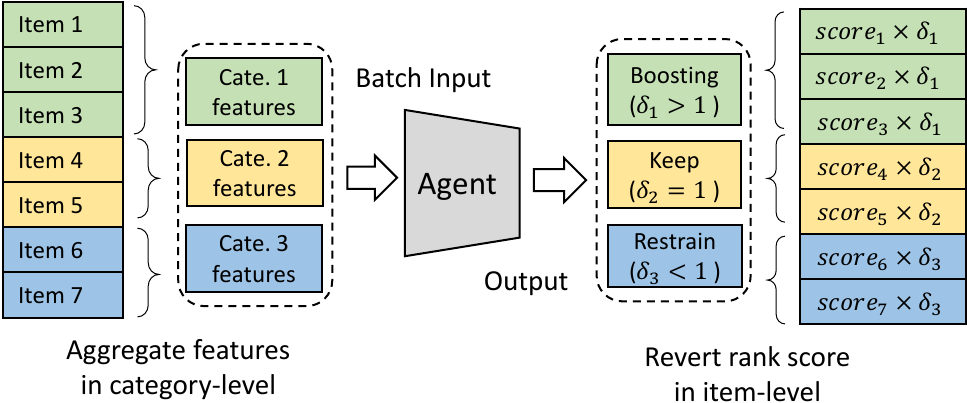}
\caption{Batch training and execution in categorical level.}
\label{fig:observation}
\end{figure}

Since we are modeling on the categorical level, we use an aggregation method to summarize the features of the items that belong to the same category as Fig. \ref{fig:observation}.
Then, the observation for each category is described by a vector of statistical features, e.g., the mean, max, min, and standard deviation of each item-level observations. 
To speed up the calculations, the agent feeds in all the categorical features of a request as a learning/execution batch, and outputs the corresponding actions. 

The recurrent model is implemented by one stack layer LSTM with the hidden size of 256, and we unroll the LSTM cell in a maximum sequence length of 25. At each time-step, the simulator outputs a 2-dim vector representing the probability of click and purchase, which are optimized by real user feedbacks. Based on the training results of the simulator, we choose several important features to the infer of a user hidden state, which contain \textit{price}, \textit{bid}, \textit{pCTR}, \textit{pv\textsubscript{gw}}, \textit{clk\textsubscript{gw}}, \textit{pCVR\textsubscript{gw}}, \textit{pv\textsubscript{gi}}, \textit{clk\textsubscript{gi}}, \textit{pCVR\textsubscript{gi}} and \textit{scen}. To work with a discrete conditional HMM, we use a quantile-based discretization for each observed feature.

\begin{table}
\tiny
\resizebox{\columnwidth}{!}{%
\begin{tabular}{c |c|c c c c }
\hline
Method & Parameter  & Revenue & Cost & ROI & Reward\\
\hline
Manual bid & - & 100\% & 100\% & 100\%& 100\% \\
\hline
\multirow{10}{*}{DISA} & 
  $\gamma$=0.1, n=5 & 100.3\% & 98.2\% & 102.2\% & 107.9\%\\
  & $\gamma$=0.3, n=5 & 111.7\% & 100.7\% & 110.9\% & 115.9\%\\
 &  $\gamma$=0.5, n=5 & 113.8\% & 101.1\% & 112.5\% & 117.5\% \\
 &  $\gamma$=0.7, n=5 & 119.0\% & 101.9\% & 116.7\% & 122.1\% \\
 &  $\gamma$=0.9, n=5 & 120.9\% & 100.7\% & 120.0\% & 125.2\%\\
 \cline{2-6}
 &  $\gamma$=0.9, n=1 & 110.6\% & 100.2\% & 110.3\% & 113.3\% \\
&  $\gamma$=0.9, n=2 & 117.1\% & 102.0\% & 114.8\% &  122.3\% \\
&  $\gamma$=0.9, n=3 & 117.6\% & 101.6\% & 115.8\% &122.7\% \\
 &  $\gamma$=0.9, n=4 & 119.5\% & 103.3\% & 115.6\% & 124.6\% \\
&  $\gamma$=0.9, n=5 & 120.9\% & 100.7\% & 120.0\% & 125.2\% \\
\hline
\end{tabular}
}
\caption{Hyper-parameter tunning in DISA}
\label{table:performance}
\end{table}

\begin{figure}
\centering
\includegraphics[height=3.0cm]{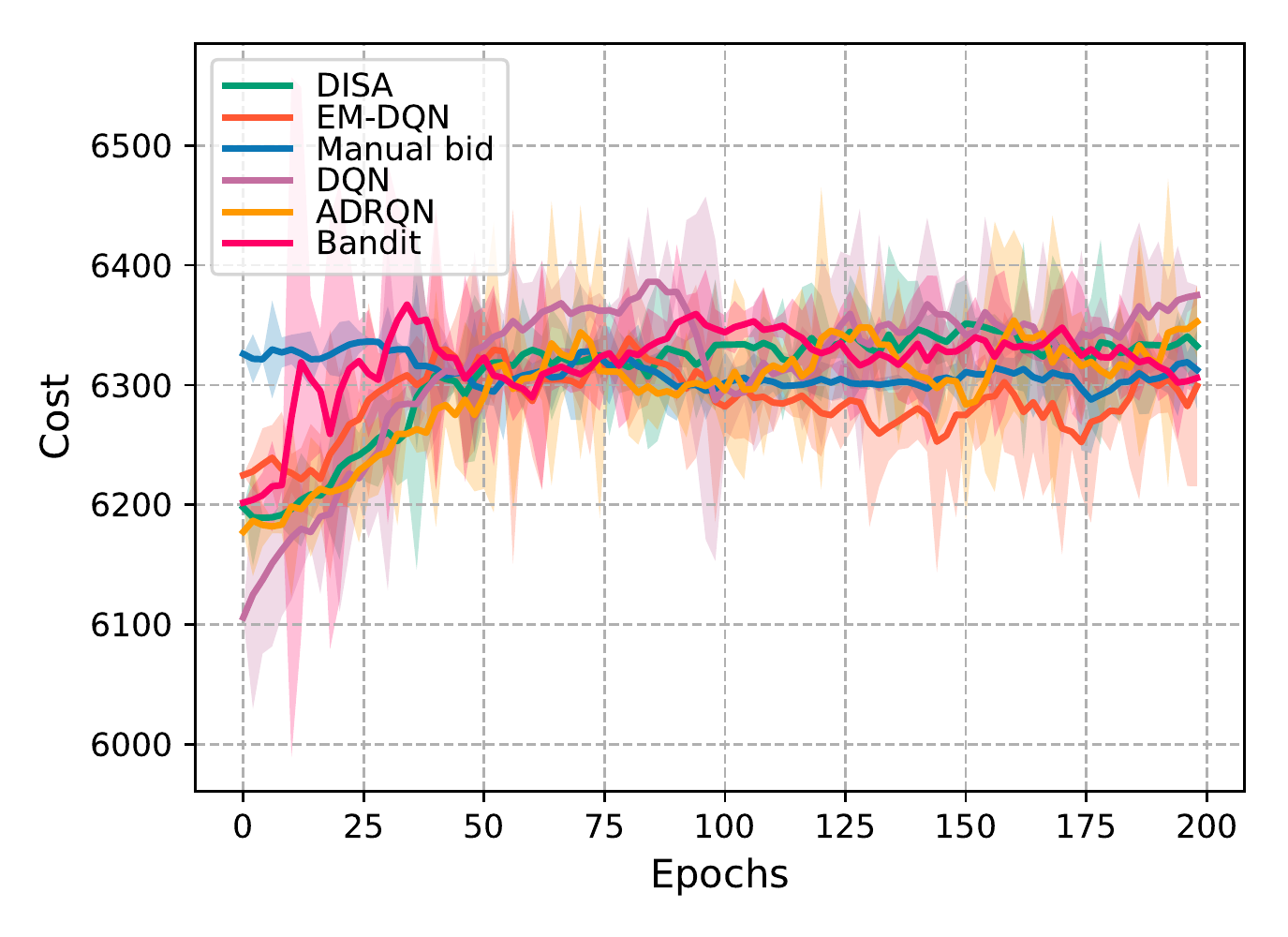}
\includegraphics[height=3.0cm]{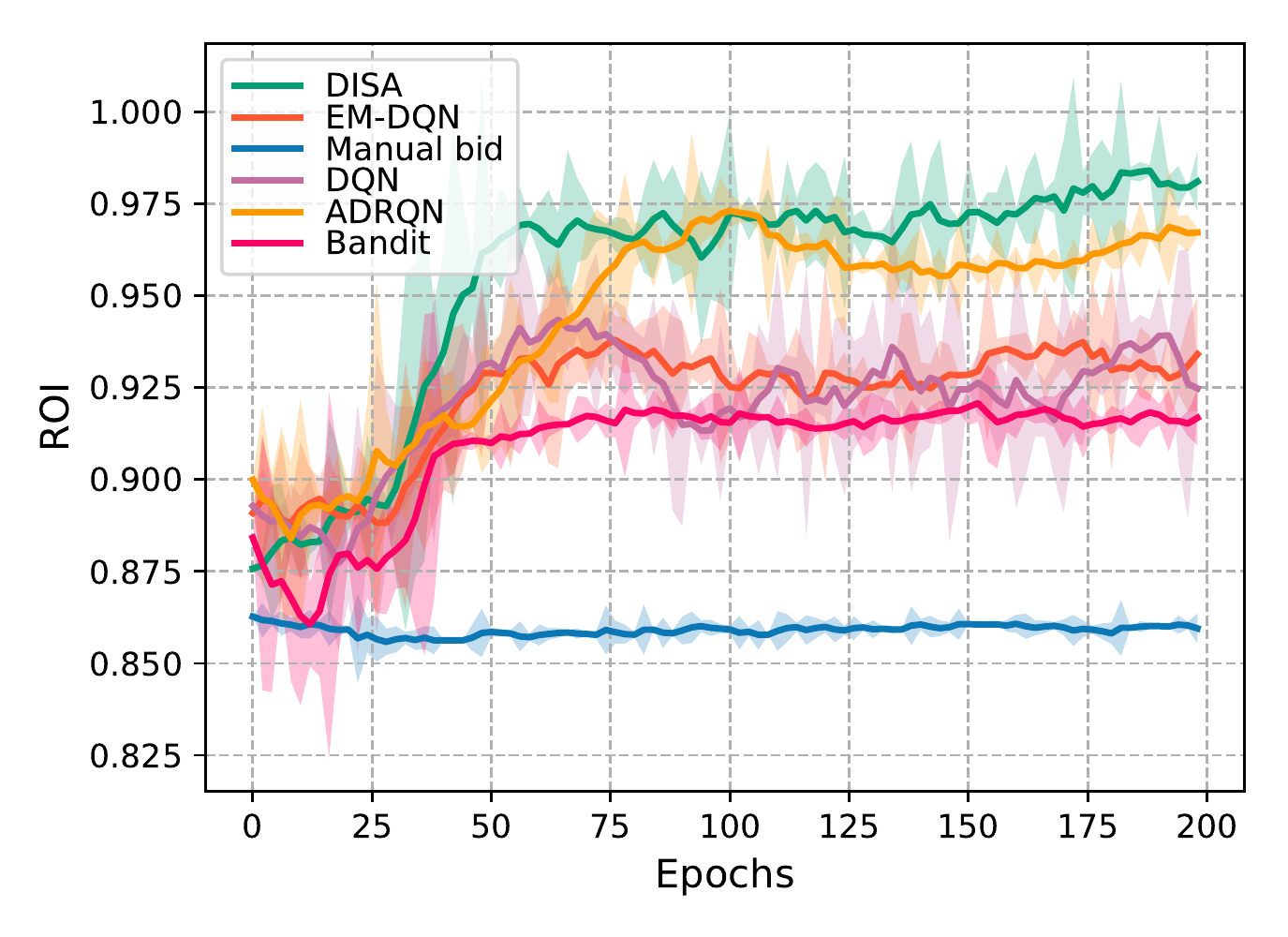}
\caption{The learning curves of cost and ROI.}
\label{fig:training}
\end{figure}

In particular, the \textit{pv\textsubscript{gi}} is discretized into a range of $[0, 6]$, and the \textit{clk\textsubscript{gi}} is discretized into a range of $[0, 5]$. 
For the \textit{pv\textsubscript{gw}}, we use the range $[0, 11]$, and \textit{clk\textsubscript{gw}} is mapped into a range of $[0, 3]$.
The \textit{scen} will equal to 1 if the current scenario is under \textit{Guess What You Like} otherwise equal to 0. 
Note that the discretization will not affect the features' monotonicity, e.g., \textit{pv\textsubscript{gw}} $= 5$ means a stronger impression being made than that of \textit{pv\textsubscript{gw}} $= 4$ in \textit{Guess What You Like}; 
$scen = 0.8$ means a higher probability of switching into \textit{Guess What You Like} than $scen = 0.5$.

The action is also discretized into three distinct values where the boosting, keeping and restraining action are defined by $\delta$=10, $\delta$=1 and $\delta$=0.1 respectively.
For an ad item, the boosting action with $\delta$=10 can almost guarantee to win the bidding, while the restraining action with $\delta$=0.1 can almost prevent its winning of the bidding.   
Since the distribution of user hidden states is stationary and will not migrate over time in our experiment, the learned parameters are fixed while optimizing the agent's policies. 

\subsubsection{Policy Learning with Trajectory Replays}
The off-policy RL is identical to our problem because the agent passively responds to user requests, and the next request might come from a different user.
Thus, for every user and category pair, we rely on a trajectory reply pool to store the corresponding experience tuple, used for constructing transition samples.
The updating of the state estimator is performed along with the policy learning to cover the patterns of newly arrived user trajectories. 
For each Q-network, the target network freezing technique is also adopted to stabilize the learning process. 
The training of DISA is formalized as Algorithm 1.


\subsubsection{Hyper-parameter Tunning} 
The discount factor $\gamma$ determines the importance of future rewards. 
In Table \ref{table:performance}, we find almost all the methods will perform better as $\gamma$ increases from 0 to 0.5. 
This result shows the existence of the future delayed rewards and proves the multi-step decision-making property of our problem. 
The value of $n$ in Eq. (9) decides how many regions the belief space will be split. 
When $n$ is 1, the belief value function for each action is represented by a linear hyperplane, and with the increase of $n$, the value function will be represented by more hyperplanes, leading to a finer and more accurate belief region.   
In our experiments, by tuning on $\gamma$ and $n$, we find the hyper-parameters of $\gamma=0.9$, $n=5$ can achieve the best performance. 
Fig. \ref{fig:training}\footnote{Each curve is smoothed on average, and the shaded area shows the standard deviation.} illustrates the learning process of different methods with the best parameter setting, from which we can see that our method DISA achieves higher ROI and also converges faster than others.


\end{document}